\documentclass[sigconf,screen]{acmart}

\AtBeginDocument{
  \fancyhead{}
  \fancyhead[LE,LO]{\small When Missing Becomes Structure: Intent-Preserving Policy Completion from Financial KOL Discourse}
  \fancyhead[RE,RO]{\small Liu, Yuan, Zhou, and Lu}
}

\setcopyright{none}

\acmConference[]{}{}{}
\acmBooktitle{}
\acmPrice{}
\acmDOI{}
\acmISBN{}

\usepackage{placeins}
\usepackage{float}
\usepackage{multirow}
\usepackage{booktabs}
\usepackage{pdflscape}
\usepackage{longtable}
\usepackage{array}
\usepackage{caption}
\usepackage{dblfloatfix}
\usepackage{tabularx}
\usepackage{cuted}
\newcolumntype{L}[1]{>{\raggedright\arraybackslash}p{#1}}
\newcolumntype{Y}{>{\raggedright\arraybackslash}X}
\usepackage{enumitem}
\providecommand{\hypersetup}[1]{}
\hypersetup{
  colorlinks=true,
  linkcolor=black,
  citecolor=blue,
  urlcolor=blue
}

\makeatletter
\@ifundefined{footnotetextcopyrightpermission}
  {\newcommand\footnotetextcopyrightpermission[1]{}}
  {\renewcommand\footnotetextcopyrightpermission[1]{}}
\makeatother

\begin{document}

\title{When Missing Becomes Structure: Intent-Preserving Policy Completion from Financial KOL Discourse}

\author{Yuncong Liu$^{1}$, Wan Yuan$^{2,*}$, Jiang Zhou$^{2,*}$, Yao Lu$^{2,\dagger}$}
\affiliation{
  \institution{National University of Singapore}
  \country{}
}
\newcommand{\authorfootnotetext}{
  $^{1}$ Asian Institute of Digital Finance, National University of Singapore; $^{2}$ School of Computing, National University of Singapore; $^{*}$ Equal contribution; $^{\dagger}$ Corresponding author.
}

\begin{abstract}
{Key Opinion Leader (KOL)} discourse on social media is widely consumed as investment guidance, yet turning it into executable trading strategies without injecting assumptions about unspecified execution decisions remains an open problem.
We observe that the gaps in KOL statements are not random deficiencies but a structured separation: KOLs express directional intent (what to buy or sell and why) while leaving execution decisions (when, how much, how long) systematically unspecified.
Building on this observation, we propose an \emph{intent-preserving policy completion} framework that treats KOL discourse as a partial trading policy and uses offline reinforcement learning to complete the missing execution decisions around the KOL-expressed intent.
Experiments on multimodal KOL discourse from YouTube and X (2022--2025) show that KICL achieves the best return and Sharpe ratio on both platforms while maintaining zero unsupported entries and zero directional reversals, and ablations confirm that the full framework yields an 18.9\% return improvement over the KOL-aligned baseline.

\end{abstract}



\maketitle
\begingroup
\renewcommand{\thefootnote}{}
\footnotetext{\authorfootnotetext}
\endgroup
\setcounter{footnote}{0}

\section{Introduction}
\begin{figure}[t]
    \centering
    \includegraphics[width=\columnwidth]{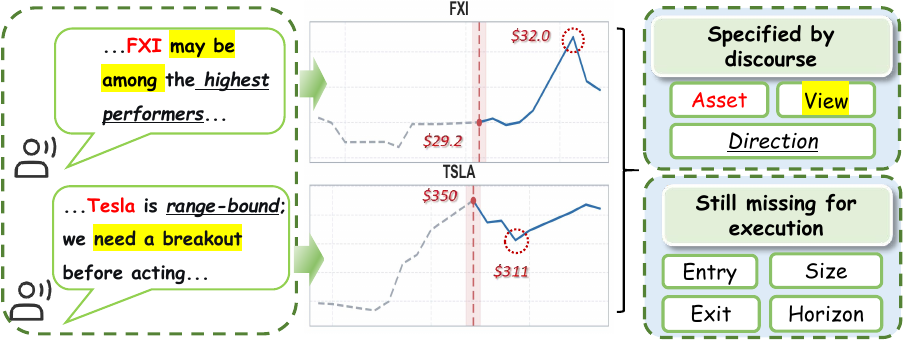}
    \caption{Representative KOL discourse examples with local price context. KOL statements specify asset-level directional intent but leave execution decisions under-specified.}
    \label{fig:kol_partial_policy_example}
\end{figure}

Financial Key Opinion Leader (KOL) discourse on social media is widely treated as a source of investment signals.
A large body of work has shown that the semantic content of financial news and online discussions is systematically related to market returns and volatility \cite{tetlock2007giving, antweiler2004all, bollen2011twitter, sprenger2014tweets, ranco2015effects, xiao2018trading}, establishing crowd-level social media sentiment as a rich information source for investment modeling.
Unlike generic social media users, financial KOLs attract sustained attention due to perceived expertise, track record, or influence over investor behavior. Their discourse is read not as background market mood, but as \emph{weighted} guidance on specific assets and directions \cite{chen2014wisdom, fang2009media, guo2015investor, kakhbod2023finfluencers}. Figure~\ref{fig:kol_partial_policy_example} illustrates this: representative KOL statements are tied to concrete market situations and convey a directional stance, even when conditional or wait-oriented, yet leave key execution decisions unspecified.

Most prior work treats social-media text as exogenous predictive signals, using crowd sentiment to forecast returns, classify price movements, or trigger trades \cite{mittal2012stock, pagolu2016sentiment, si2013exploiting, sprenger2014tweets, schumaker2009textual, ke2019predicting, ding2015deep}. Financial KOL discourse is fundamentally different. As Table~\ref{tab:kol_discourse_stats_main} shows, it is highly asset-specific and directional (ticker-explicit ratio $>$79\% on both platforms; directional-expression ratio $>$79\%), yet execution-complete descriptions are almost nonexistent (full-execution completeness: 1.86\% on YouTube, 0.00\% on X). Follow-up is equally sparse: over 65\% of mentioned assets receive no follow-up within 90~days. This combination of strong directional specificity with near-absent execution completeness indicates that KOL discourse functions less like generic sentiment and more like a \emph{partially specified basis for decision-making}.

\vspace{0.05in}\noindent\textbf{Structural Incompleteness of KOL Discourse}.
Table~\ref{tab:kol_discourse_stats_main} confirms across 18 YouTube KOLs and 14 X KOLs that discourse is usually asset-specific and directional, but explicit entry, sizing, holding, and exit information is rare, and fully specified execution is almost never observed. This incompleteness is not random noise but a stable structural property of how KOL intent is expressed.

\vspace{0.05in}\noindent\textbf{From Partial Intent to Policy Completion}.
Figure~\ref{fig:kol_partial_policy_example} makes the missing part concrete. The X-side example (``\$FXI may be among the highest performers'') identifies the asset and conveys a positive relative view; the YouTube-side example (``Tesla is range-bound; we need a breakout before acting'') specifies a conditional wait-oriented stance. Neither statement tells us \emph{how} to trade the view through time: no entry rule, no sizing, no holding horizon, no exit plan. We therefore interpret financial KOL discourse as specifying a \emph{partial trading policy} \cite{parr1997reinforcement, jaakkola1994reinforcement, liu2022partially}: it expresses directional intent but leaves execution-level decisions unspecified.

This reinterpretation changes what must be learned. The problem is not to infer an entire trading strategy from scratch, but to \emph{complete} the missing execution components around KOL-specified intent. We therefore formulate learning from financial KOL discourse as intent-preserving policy completion rather than unconstrained strategy generation. This work makes the following contributions:
\begin{itemize}[leftmargin=0.15in]
  \item We identify a structural property of financial KOL discourse: its incompleteness reflects a systematic pattern in how investment intent is expressed, rather than noise or missing information. This observation reveals that KOL discourse naturally specifies only a subset of trading decisions, which can be interpreted as a partial trading policy.

  \item Building on this insight, we propose \textbf{KICL} (KOL Intent Constrained Learning), an intent-preserving policy completion framework that formulates learning from KOL discourse as an offline sequential decision-making problem, where reinforcement learning completes missing execution decisions while preserving the original KOL intent.
  
  \item We further introduce a betrayal-oriented evaluation perspective for KOL-conditioned policy learning, including metrics that quantify unsupported entries, directional reversals, and deviations from KOL-aligned intent.
\end{itemize}

Experiments on multimodal KOL discourse from YouTube and X (2022--2025) show that KICL achieves the best return and Sharpe ratio on both platforms while driving unsupported entry rate and directional reversal rate to zero. Ablations confirm that the full framework yields an 18.9\% return improvement and a 1.1\% Sharpe improvement over the KOL-aligned baseline, and that removing hard constraints causes a 65.8\% return collapse. Together, these results demonstrate a principled path from partially specified KOL discourse to executable, intent-preserving trading policies.

\begin{table}[t]
\centering
\caption{Empirical evidence that financial KOL discourse behaves like a partial trading policy on the full dataset (18 YouTube KOLs and 14 X KOLs).}
\label{tab:kol_discourse_stats_main}
\small
\setlength{\tabcolsep}{4pt}
\begin{tabular}{lcc}
\toprule
\textbf{Statistic} & \textbf{YouTube} & \textbf{X} \\
\midrule
\multicolumn{3}{l}{\textit{Coverage / specificity}} \\
Mentioned companies & 6,774 & 3,811 \\

Ticker-explicit ratio & 81.61\% & 79.51\% \\
Directional-expression ratio & 79.58\% & 91.62\% \\
Single-mention ratio & 62.06\% & 39.67\% \\
\midrule
\multicolumn{3}{l}{\textit{Execution under-specification}} \\
Explicit entry ratio & 33.14\% & 5.48\% \\
Explicit sizing ratio & 12.48\% & 0.62\% \\
Explicit exit ratio & 14.59\% & 2.08\% \\
Full-execution completeness ratio & 1.86\% & 0.00\% \\
\midrule
\multicolumn{3}{l}{\textit{Temporal maintenance}} \\
No follow-up within 90 days & 82.88\% & 65.81\% \\
Median time to first reversal & 179.84 days & 161.00 days \\
\bottomrule
\end{tabular}
\end{table}

\section{Related Work}
\vspace{0.05in}\noindent\textbf{Social Media Signals in Financial Markets}.
The link between textual information and financial markets is well established.
Tetlock \cite{tetlock2007giving} shows that pessimistic news language predicts downward market movements; Antweiler and Frank \cite{antweiler2004all} find that message-board discussions carry signals related to returns and volatility.
Subsequent work extends this to social media: Bollen et al.~\cite{bollen2011twitter} report correlations between Twitter mood and stock indices, Sprenger et al.~\cite{sprenger2014tweets} link StockTwits sentiment to abnormal trading activity, and further studies explore predictive value for returns and volatility \cite{ranco2015effects, mittal2012stock}.
Across this literature, language is treated as an exogenous predictor rather than as a structured basis in decision-making.

\vspace{0.05in}\noindent\textbf{Financial KOLs and Influencer-Driven Investment Signals}.
Financial KOLs and online influencers play a distinct role in shaping investor attention and decisions.
Guo et al.~\cite{guo2015investor} show that social-media information influences investor attention and trading activity; Kakhbod et al.~\cite{kakhbod2023finfluencers} document how finfluencers shape investor beliefs and market participation.
In traditional finance, Womack \cite{womack1996brokerage} finds that analyst recommendations generate significant price reactions, and Barber et al.~\cite{barber2001can} show that recommendation-based portfolios produce meaningful returns.
More recently, Hii and Ong \cite{hii2026finfluencer} show that influencer credibility affects followers’ financial decisions, and Hayes and Ben-Shmuel \cite{hayes2024under} analyze how finfluencers reshape public market interpretations.
These findings indicate that KOL discourse is consumed as actionable guidance tied to identifiable speakers---a more \emph{weighted}, actor-specific form of market guidance than generic crowd sentiment, even though it rarely specifies a complete executable strategy.

\begin{figure*}[t]
    \centering
    \includegraphics[width=\textwidth]{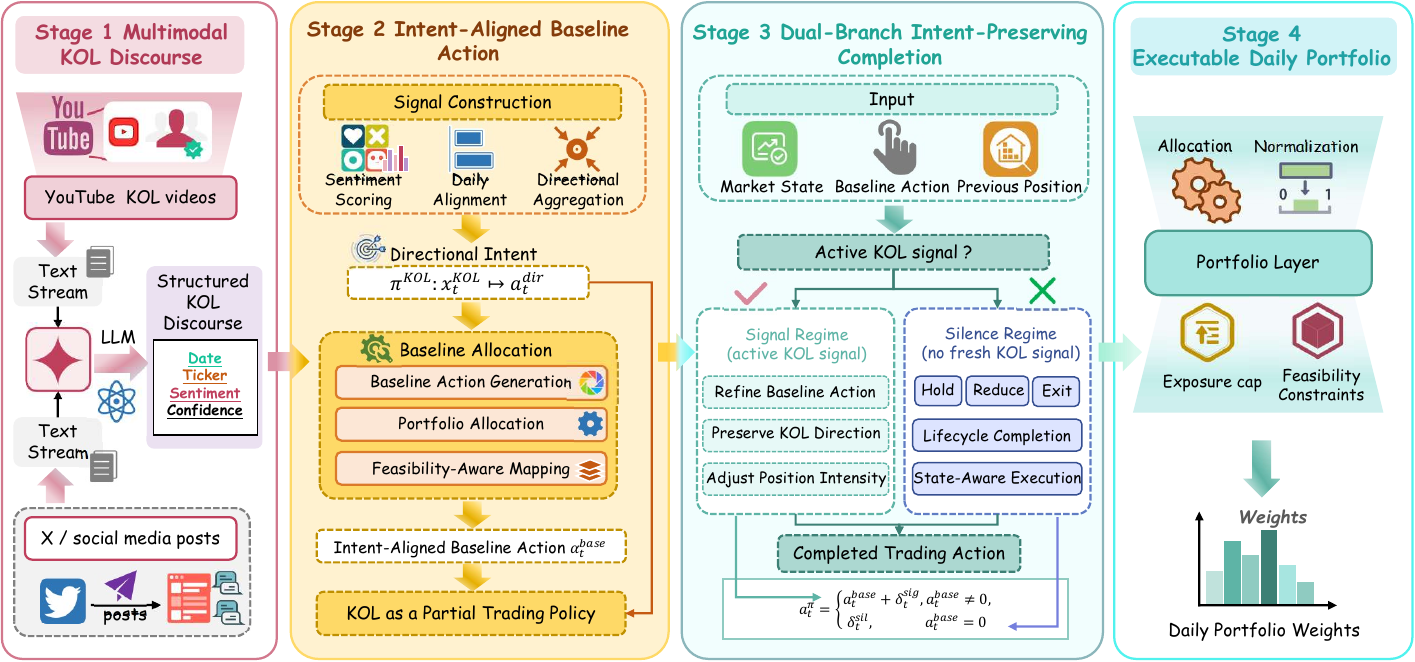}
    \caption{Overview of our intent-preserving policy completion framework. Raw financial KOL discourse is transformed into structured signals and mapped to an intent-aligned baseline action. A signal-conditioned completion policy refines the baseline under active KOL signals and completes missing execution behavior. The final action is mapped to executable portfolio weights.}
    \label{fig:framework_overview}
\end{figure*}

\vspace{0.05in}\noindent\textbf{Text-Based Trading and Signal-Driven Strategies}.
A large body of work converts language into sentiment scores or rule-based buy--sell signals \cite{sprenger2014tweets, mittal2012stock, si2013exploiting, pagolu2016sentiment}, demonstrating that online opinions can become actionable signals.
However, executable trading strategies require sequential, state-dependent decisions---position sizing, holding duration, exit timing \cite{moody2001learning, nevmyvaka2006reinforcement, jiang2017deep}---that signal-driven formulations do not specify, leaving policy construction dependent on additional assumptions about execution \cite{sutton1999between, hadfield2016cooperative}.

\vspace{0.05in}\noindent\textbf{Reinforcement Learning for Trading}.
Trading and portfolio management have been widely formulated as RL problems \cite{moody2001learning, nevmyvaka2006reinforcement, jiang2017deep, deng2016deep, yang2020deep}, and recent advances in offline RL enable learning from fixed historical datasets \cite{kostrikov2021offline}.
Existing RL-based approaches learn policies directly from prices, technical indicators, or portfolio states, while language-based approaches remain at the signal level. How to combine sequential policy learning with partially specified, actor-weighted financial discourse has received little attention.

\section{Problem Formulation}
Let the trading action at time $t$ be
$a_t = (a_t^{\mathrm{dir}}, a_t^{\mathrm{exec}})$,
where $a_t^{\mathrm{dir}}$ denotes directional intent and $a_t^{\mathrm{exec}}$ denotes execution-level decisions such as sizing, holding, and exit.
Given the empirical characteristics above, we model KOL discourse as specifying primarily the directional component,
\begin{equation}
\pi^{\mathrm{KOL}} : x_t^{\mathrm{KOL}} \mapsto a_t^{\mathrm{dir}},
\end{equation}
while leaving most execution details unspecified.
KOL discourse thus does not define a complete executable trading policy over the full horizon, but a partially specified one.
The objective is therefore to construct a full policy
\begin{equation}
\pi : s_t \mapsto (a_t^{\mathrm{dir}}, a_t^{\mathrm{exec}})
\end{equation}
that preserves the directional intent expressed by $\pi^{\mathrm{KOL}}$ while completing the missing execution decisions.

\vspace{0.05in}\noindent\textbf{Intent-Preserving Policy Completion}.
To operationalize this sparse directional intent in practice, we introduce a KOL-aligned baseline action \(a_t^{\mathrm{base}}\), which serves as an executable proxy of the partial policy.
Figure~\ref{fig:framework_overview} provides an overview of the proposed framework that operationalizes this formulation.

Let $a_t^{\mathrm{base}}$ denote the KOL-aligned baseline action.
We partition the replay buffer into signal and silence subsets:
\begin{equation}
D_{\mathrm{sig}} = \{(s_t,a_t,r_t,s_{t+1}) : a_t^{\mathrm{base}} \neq 0\},
\end{equation}
\begin{equation}
D_{\mathrm{sil}} = \{(s_t,a_t,r_t,s_{t+1}) : a_t^{\mathrm{base}} = 0\}.
\end{equation}
Here, $D_{\mathrm{sig}}$ contains transitions with an active KOL-aligned signal, while $D_{\mathrm{sil}}$ contains silence-regime transitions without fresh KOL guidance.
Empirically, the offline dataset is strongly silence-dominated:
\begin{equation}
|D_{\mathrm{sig}}| = 45{,}626,\qquad
|D_{\mathrm{sil}}| = 1{,}183{,}395,\qquad
\frac{|D_{\mathrm{sil}}|}{|D_{\mathrm{sig}}|} = 25.93,
\end{equation}
and 99.73\% of tickers satisfy $|D_{\mathrm{sil}}^{(i)}| > |D_{\mathrm{sig}}^{(i)}|$.
We therefore treat
\begin{equation}
|D_{\mathrm{sil}}| \gg |D_{\mathrm{sig}}|
\label{eq:silence_domination}
\end{equation}
as an empirical structural property of the replay buffer. 
For a generic unconstrained empirical objective
\begin{equation}
\mathcal{L}_{\mathrm{RL}}(\theta)=\frac{1}{|D|}\sum_{z\in D}\ell_\theta(z),
\end{equation}
the signal--silence decomposition gives
\begin{equation}
\mathcal{L}_{\mathrm{RL}}(\theta)
=
\frac{|D_{\mathrm{sig}}|}{|D|}\mathcal{L}_{\mathrm{sig}}(\theta)
+
\frac{|D_{\mathrm{sil}}|}{|D|}\mathcal{L}_{\mathrm{sil}}(\theta),
\end{equation}
and therefore
\begin{equation}
\nabla_\theta \mathcal{L}_{\mathrm{RL}}(\theta)
=
\frac{|D_{\mathrm{sig}}|}{|D|}\nabla_\theta \mathcal{L}_{\mathrm{sig}}(\theta)
+
\frac{|D_{\mathrm{sil}}|}{|D|}\nabla_\theta \mathcal{L}_{\mathrm{sil}}(\theta).
\end{equation}
Under Eq.~\eqref{eq:silence_domination}, unconstrained optimization is dominated by silence-regime transitions unless signal-step gradients are disproportionately larger.
Because $D_{\mathrm{sil}}$ carries no fresh KOL-aligned signal, these updates are driven by market features rather than KOL intent, shifting optimization away from discourse supervision.
Return maximization alone therefore does not guarantee intent preservation.

Let \(a^{\mathrm{base}}(s)\) denote the KOL-aligned baseline action at state \(s\). On states with active KOL signals, an unconstrained policy selects
\begin{equation}
a^{\mathrm{free}}(s)\in \arg\max_a Q(s,a),
\label{eq:free_action_selection}
\end{equation}
so whenever there exists an action \(a \neq a^{\mathrm{base}}(s)\) such that
\begin{equation}
Q(s,a) > Q(s,a^{\mathrm{base}}(s)),
\label{eq:free_deviation_condition}
\end{equation}
the learned policy will deviate from the KOL-aligned action at that state.
KOL intent is therefore not guaranteed to be preserved under unrestricted return-driven optimization.

The problem is thus not unrestricted policy optimization, but intent-preserving completion of a partially specified policy. Let
\begin{equation}
m_t = \mathbf{1}\!\left[a_t^{\mathrm{base}} \neq 0\right],
\qquad
p_{t-1} = \operatorname{sign}(a_{t-1}),
\end{equation}
where \(m_t\) indicates whether an active KOL-aligned signal is present and \(p_{t-1}\) denotes the direction of the carried position from the previous step. We define the admissible directional action set as
\begin{equation}
\mathcal{A}_t^{\mathrm{dir}} =
\begin{cases}
\left\{\operatorname{sign}(a_t^{\mathrm{base}})\right\}, & m_t = 1, \\[4pt]
\{0\}, & m_t = 0,\; p_{t-1}=0, \\[4pt]
\left\{0,\; p_{t-1}\right\}, & m_t = 0,\; p_{t-1}\neq 0 .
\end{cases}
\label{eq:feasible_dir_set}
\end{equation}
The constrained policy completion objective is then
\begin{equation}
\pi^\ast
=
\arg\max_{\pi}\; \mathbb{E}[R(\pi)]
\quad
\text{s.t.}
\quad
a_t^{\mathrm{dir}} \in \mathcal{A}_t^{\mathrm{dir}},\ \forall t .
\label{eq:intent_constraint}
\end{equation}
Under this formulation, the policy is free to complete execution-level decisions such as sizing and exit timing within the admissible directional set, while unsupported entries and directional reversals are ruled out by construction.

\FloatBarrier

\section{Methodology}

\begin{figure*}[t]
    \centering
    \includegraphics[width=\textwidth]{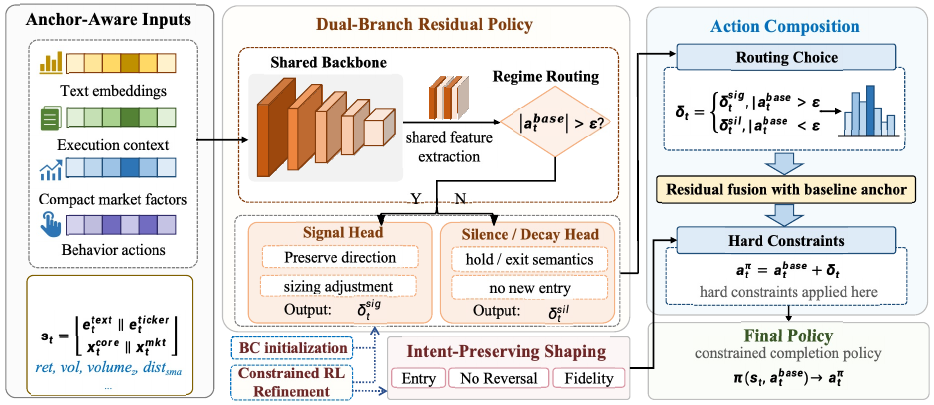}
    \caption{Overview of KICL. Anchor-aware inputs are mapped to a dual-branch residual policy with regime routing. The chosen residual is fused with the KOL-aligned baseline and filtered by hard constraints to produce the final action.}
    \label{fig:kicl_overview}
\end{figure*}

\noindent\textbf{Overview}.
We propose \textbf{KICL} (KOL-Intent-Constrained Learning), a baseline-anchored policy completion framework for constructing executable trading policies from financial KOL discourse. Figure~\ref{fig:kicl_overview} summarizes the architecture.

\begin{equation}
\pi : s_t \mapsto \left(a_t^{\mathrm{dir}}, a_t^{\mathrm{exec}}\right),
\end{equation}

\noindent where KOL discourse specifies the baseline decision skeleton and offline reinforcement learning completes the under-specified execution dynamics. As shown in Figure~\ref{fig:framework_overview}, the framework has four stages: structured KOL signal construction, baseline action generation, signal-conditioned policy completion, and executable portfolio mapping.

\vspace{0.05in}\noindent\textbf{Structured KOL Signals and Baseline Policy}.
Raw KOL discourse is first mapped to a structured signal representation aligned with time and asset identity:
\begin{equation}
x_t^{\mathrm{KOL}} = (t,\ \mathrm{ticker},\ \mathrm{sentiment},\ \mathrm{confidence}),
\end{equation}
where \textit{sentiment} and \textit{confidence} encode the polarity and strength of the expressed opinion.
Given $x_t^{\mathrm{KOL}}$, we construct an intent-aligned baseline action
\begin{equation}
a_t^{\mathrm{base}} = f_{\mathrm{base}}\!\left(x_t^{\mathrm{KOL}}\right),
\end{equation}
which maps the extracted signal into a baseline action score derived from sentiment and confidence. This score is then converted into executable anchor weights
\begin{equation}
w_t^{\mathrm{base}} = \mathrm{Alloc}\!\left(a_t^{\mathrm{base}}\right),
\end{equation}
where $\mathrm{Alloc}(\cdot)$ denotes the constrained portfolio layer. In practice, this executable baseline serves as the intent anchor throughout replay construction, policy completion, and evaluation. It both operationalizes sparse KOL intent as a tradable proxy and defines the action reference around which completion is constrained.

\vspace{0.05in}\noindent\textbf{Policy Completion under Signal and Silence Regimes}.
To complete the missing execution component, we define the policy on an augmented state
\begin{equation}
s_t = \bigl(s_t^{\mathrm{mkt}},\ a_t^{\mathrm{base}},\ p_{t-1},\ \tau_t \bigr),
\end{equation}
where $s_t^{\mathrm{mkt}}$ denotes market features, $a_t^{\mathrm{base}}$ the KOL-aligned baseline action, $p_{t-1}$ the previous position, and $\tau_t$ the elapsed time since the last explicit KOL signal.
The completion policy is conditioned on whether the baseline action is active.
When $a_t^{\mathrm{base}}\neq 0$,
\begin{equation}
a_t^{\pi,\mathrm{sig}} = a_t^{\mathrm{base}} + \delta_t^{\mathrm{sig}},
\qquad
\delta_t^{\mathrm{sig}} = h_\theta\!\left(s_t^{\mathrm{mkt}},\ a_t^{\mathrm{base}}\right),
\end{equation}
so the policy refines the KOL-aligned draft rather than replacing it.
When $a_t^{\mathrm{base}}=0$,
\begin{equation}
a_t^{\pi,\mathrm{sil}} = \delta_t^{\mathrm{sil}},
\qquad
\delta_t^{\mathrm{sil}} = g_\theta\!\left(s_t^{\mathrm{mkt}},\ p_{t-1},\ \tau_t\right),
\end{equation}
so the policy completes the unresolved execution lifecycle of prior KOL-induced positions (hold, reduction, exit) rather than inferring unsupported new intent.
The final policy is therefore
\begin{equation}
a_t^\pi =
\begin{cases}
a_t^{\mathrm{base}} + \delta_t^{\mathrm{sig}}, & a_t^{\mathrm{base}} \neq 0,\\[4pt]
\delta_t^{\mathrm{sil}}, & a_t^{\mathrm{base}} = 0.
\end{cases}
\end{equation}
Thus, the policy remains baseline-anchored under active signals and transitions to lifecycle-oriented execution completion under silence.

\vspace{0.05in}\noindent\textbf{Offline Optimization with Intent Constraints}. 
We learn the policy from a fixed historical dataset
\begin{equation}
\mathcal D=\{(s_t,a_t,r_t,s_{t+1})\}_{t=1}^{N},
\end{equation}
without online interaction.
Our objective is not unrestricted policy optimization, but baseline-anchored policy completion:
\begin{equation}
\pi^\ast
=
\arg\min_{\pi}\ \mathcal L_{\mathrm{total}}
\quad
\text{s.t.}\quad
a_t^\pi =
\begin{cases}
a_t^{\mathrm{base}} + \delta_t^{\mathrm{sig}}, & a_t^{\mathrm{base}} \neq 0,\\[4pt]
\delta_t^{\mathrm{sil}}, & a_t^{\mathrm{base}} = 0.
\end{cases}
\end{equation}

For the offline RL backbone, we adopt a data-supported value-learning objective in the spirit of Implicit Q-Learning (IQL) \cite{kostrikov2021offline}, followed by advantage-weighted policy extraction over dataset-supported actions.
Dataset support alone is insufficient in our setting, however.
As shown in Eqs.~\eqref{eq:free_action_selection} and \eqref{eq:free_deviation_condition}, unconstrained optimization selects
\begin{equation}
a^{\mathrm{free}}(s)\in \arg\max_a Q(s,a),
\end{equation}
so any action $a \neq a^{\mathrm{base}}(s)$ will be preferred whenever
\begin{equation}
Q(s,a) > Q(s,a^{\mathrm{base}}(s)),
\end{equation}
which permits market-driven deviations from the KOL-aligned baseline.
To prevent this, we optimize the completion policy under
\begin{equation}
\mathcal L_{\mathrm{total}}
=
\mathcal L_{\mathrm{RL}}
+
\lambda_{\mathrm{fid}}\mathcal L_{\mathrm{fid}}
+
\lambda_{\mathrm{rev}}\mathcal L_{\mathrm{rev}}
+
\lambda_{\mathrm{ent}}\mathcal L_{\mathrm{ent}}.
\end{equation}
Here, $\mathcal L_{\mathrm{fid}}$ limits overall drift from the baseline, while $\mathcal L_{\mathrm{rev}}$ and $\mathcal L_{\mathrm{ent}}$ suppress baseline-opposite reversals and unsupported new entries under silence.
The constrained objective induces a baseline-aware score:
\begin{equation}
\begin{aligned}
\widetilde Q(s,a)
={}&\, Q(s,a)
-\lambda_{\mathrm{fid}}\, d_{\mathrm{fid}}\!\left(a, a^{\mathrm{base}}(s)\right) \\
&-\lambda_{\mathrm{rev}}\, d_{\mathrm{rev}}\!\left(a, a^{\mathrm{base}}(s)\right) \\
&-\lambda_{\mathrm{ent}}\, d_{\mathrm{ent}}\!\left(a, a^{\mathrm{base}}(s)\right).
\end{aligned}
\end{equation}
and the selected action becomes
\begin{equation}
a^{\mathrm{ours}}(s)\in \arg\max_a \widetilde Q(s,a).
\end{equation}

\noindent We define the total deviation barrier as
\begin{equation}
\begin{aligned}
\Delta_{\mathrm{dev}}(s,a)
={}&\, \lambda_{\mathrm{fid}}\, d_{\mathrm{fid}}\!\left(a, a^{\mathrm{base}}(s)\right) \\
&+\lambda_{\mathrm{rev}}\, d_{\mathrm{rev}}\!\left(a, a^{\mathrm{base}}(s)\right) \\
&+\lambda_{\mathrm{ent}}\, d_{\mathrm{ent}}\!\left(a, a^{\mathrm{base}}(s)\right).
\end{aligned}
\end{equation}
A deviating action is admitted only if
\begin{equation}
Q(s,a)-Q\!\left(s,a^{\mathrm{base}}(s)\right)>\Delta_{\mathrm{dev}}(s,a).
\label{eq:constrained_deviation_condition}
\end{equation}
Compared with Eq.~\eqref{eq:free_deviation_condition}, deviation now requires overcoming an explicit intent-preservation barrier rather than relying on return advantage alone.

\begin{table}[t]
\centering
\caption{Evaluation metrics used in experiments.}
\label{tab:eval_metrics}
\small
\setlength{\tabcolsep}{3pt}
\renewcommand{\arraystretch}{1.02}
\begin{tabular}{@{}p{1.15cm}p{6.1cm}@{}}
\toprule
\textbf{Metric} & \textbf{Description} \\
\midrule
\multicolumn{2}{@{}l}{\textit{Performance Metrics}} \\
\textbf{Return} & Cumulative return over the evaluation period \\
\textbf{Sharpe} & Risk-adjusted return based on daily returns \\
\textbf{MDD} & Largest peak-to-trough drawdown \\
\textbf{WR} & Event-level win rate against the executable KOL-aligned baseline \\
\midrule
\multicolumn{2}{@{}l}{\textit{Betrayal Metrics}} \\
\textbf{UER} & Fraction of entries without an active KOL baseline \\
\textbf{DRR} & Fraction of active-signal steps opposing the baseline \\
\textbf{BD} & Mean absolute policy--baseline action deviation \\
\textbf{CG} & Same-direction co-movement mismatch to the baseline \\
\bottomrule
\end{tabular}
\end{table}

\begin{table}[t]
\centering
\caption{Summary statistics of the constructed dataset.}
\label{tab:dataset_summary}
\small
\setlength{\tabcolsep}{4pt}
\begin{tabular}{lcc}
\toprule
\textbf{Statistic} & \textbf{YouTube} & \textbf{X} \\
\midrule
\multicolumn{3}{l}{\textit{Source scope}} \\
KOLs (full / benchmark) & 18 / 10 & 14 / 10 \\
Collection period & \multicolumn{2}{c}{Jan 2022 -- Mar 2025} \\
Mentioned companies & 6,774 & 3,811 \\
\midrule
\multicolumn{3}{l}{\textit{Signal characteristics}} \\
Ticker-explicit ratio & 81.61\% & 79.51\% \\
Directional-expression ratio & 79.58\% & 91.62\% \\
Full-execution completeness & 1.86\% & 0.00\% \\
\midrule
\multicolumn{3}{l}{\textit{Trajectory statistics}} \\
Total transitions & \multicolumn{2}{c}{1,229,021} \\
Signal transitions ($|D_{\mathrm{sig}}|$) & \multicolumn{2}{c}{45,626 \;(3.71\%)} \\
Silence transitions ($|D_{\mathrm{sil}}|$) & \multicolumn{2}{c}{1,183,395 \;(96.29\%)} \\
Silence-to-signal ratio & \multicolumn{2}{c}{25.93} \\
\bottomrule
\end{tabular}
\end{table}

\begin{table*}[!t]
\centering
\caption{Main benchmark results on the selected 20-KOL subset under the upgraded event-conditioned benchmark. Methods are grouped by family. Best values are shown in bold and second-best values are underlined within each platform.}\label{tab:main_results}
\small
\setlength{\tabcolsep}{3.8pt}
\renewcommand{\arraystretch}{1.12}
\begin{tabular}{@{}llccccccc|ccccccc@{}}
\toprule
& & \multicolumn{7}{c|}{\textbf{X}} & \multicolumn{7}{c}{\textbf{YouTube}} \\
\cmidrule(lr){3-9}\cmidrule(l){10-16}
& & \multicolumn{4}{c}{\textit{Performance}} & \multicolumn{3}{c|}{\textit{Betrayal}} & \multicolumn{4}{c}{\textit{Performance}} & \multicolumn{3}{c}{\textit{Betrayal}} \\
\cmidrule(lr){3-6}\cmidrule(lr){7-9}\cmidrule(lr){10-13}\cmidrule(l){14-16}
\textbf{Group} & \textbf{Method}
& \textbf{Ret}$\uparrow$ & \textbf{Sha}$\uparrow$ & \textbf{MDD}$\downarrow$ & \textbf{WR}$\uparrow$
& \textbf{UER}$\downarrow$ & \textbf{DRR}$\downarrow$ & \textbf{BD}$\downarrow$
& \textbf{Ret}$\uparrow$ & \textbf{Sha}$\uparrow$ & \textbf{MDD}$\downarrow$ & \textbf{WR}$\uparrow$
& \textbf{UER}$\downarrow$ & \textbf{DRR}$\downarrow$ & \textbf{BD}$\downarrow$ \\
\midrule

\multirow{2}{*}{Heuristic}
& RMB
& 0.02 & 0.30 & 0.33 & \underline{0.70}
& \textbf{0.00} & \textbf{0.00} & \textbf{0.00}
& 0.27 & \underline{2.52} & 0.16 & 0.50
& \textbf{0.00} & \textbf{0.00} & \textbf{0.00} \\
& HAP
& 0.00 & -0.07 & 0.11 & \underline{0.70}
& \underline{0.19} & 0.19 & 0.02
& 0.12 & 2.43 & 0.07 & 0.30
& 0.35 & \underline{0.12} & 0.04 \\
\midrule

\multirow{2}{*}{Imit./Sup.}
& BC~\cite{bain1995framework}
& 0.01 & -0.02 & 0.11 & \textbf{0.80}
& 0.22 & \underline{0.19} & 0.02
& 0.09 & 1.94 & 0.07 & 0.30
& 0.27 & 0.24 & 0.04 \\
& SUP-DELTA
& 0.02 & \underline{0.30} & 0.32 & 0.40
& \textbf{0.00} & \textbf{0.00} & 0.01
& \underline{0.29} & 2.47 & 0.16 & \underline{0.70}
& \textbf{0.00} & \textbf{0.00} & 0.01 \\
\midrule

\multirow{4}{*}{Offline RL}
& IQL~\cite{kostrikov2021offline}
& 0.01 & -0.03 & 0.10 & \textbf{0.80}
& 0.19 & 0.21 & 0.03
& 0.12 & 2.43 & 0.08 & 0.30
& \underline{0.25} & 0.19 & 0.03 \\
& AWAC~\cite{ashvin2020accelerating}
& \underline{0.03} & 0.06 & 0.10 & \underline{0.70}
& 0.19 & 0.19 & 0.03
& 0.12 & 2.41 & 0.07 & 0.40
& 0.27 & 0.18 & 0.04 \\
& CQL~\cite{kumar2020conservative}
& -0.01 & -0.09 & \underline{0.09} & 0.50
& 0.84 & 0.38 & 0.10
& 0.01 & 0.51 & \textbf{0.05} & 0.10
& 0.92 & 0.33 & 0.24 \\
& TD3+BC~\cite{fujimoto2021minimalist}
& -0.02 & -0.39 & \textbf{0.08} & 0.30
& 0.79 & 0.37 & 0.18
& -0.01 & 0.03 & \underline{0.06} & 0.10
& 0.94 & 0.50 & 0.21 \\
\midrule

Ours
& \textbf{KICL}
& \textbf{0.11} & \textbf{0.47} & 0.32 & \textbf{0.80}
& \textbf{0.00} & \textbf{0.00} & \underline{0.01}
& \textbf{0.29} & \textbf{2.56} & 0.16 & \textbf{0.90}
& \textbf{0.00} & \textbf{0.00} & \underline{0.01} \\
\bottomrule
\end{tabular}
\end{table*}

\section{Experiments}
\subsection{Dataset Construction}
We construct an intent-aligned offline benchmark from multimodal financial KOL discourse on YouTube and X, spanning January 2022 to March 2025. The benchmark is built through source collection, structured signal extraction, market-data alignment, and trajectory construction.

\vspace{0.05in}\noindent\textbf{Source Collection}.
We collect public discourse from 18 YouTube KOLs and 14 X KOLs selected for sustained activity, identifiable asset-level commentary, and audience reach.
YouTube data comprises full video transcripts (auto-generated and manually corrected where available), video titles, and upload metadata; X data comprises post text, timestamps, and referenced tickers.
The two platforms exhibit complementary discourse characteristics: YouTube KOLs tend to produce longer, structured analyses with conditional reasoning (conditional-action ratio 47.19\%), while X KOLs produce shorter, more directionally explicit statements (directional-expression ratio 91.62\%), as reported in Table~\ref{tab:kol_discourse_stats_main}.

\vspace{0.05in}\noindent\textbf{Structured Signal Extraction}.
Raw multimodal discourse is mapped to a structured signal representation aligned with time and asset identity:
\begin{equation}
x_t^{\mathrm{KOL}} = (t,\ \mathrm{ticker},\ \mathrm{sentiment},\ \mathrm{confidence}).
\end{equation}
For YouTube, we first transcribe video content and segment it into asset-level discussion blocks using temporal and topic cues. For X, individual posts are directly parsed for ticker mentions and directional language.
In both cases, a large language model extracts sentiment polarity (positive, negative, neutral) and a confidence score reflecting the expressed conviction strength.
Ticker resolution maps informal references (e.g., ``Tesla'', ``TSLA'', ``\$TSLA'') to canonical identifiers. Signals that cannot be resolved to a specific tradable asset are discarded.

\vspace{0.05in}\noindent\textbf{Market Data Alignment}.
Each extracted KOL signal is aligned with daily market data for the referenced asset. The market state $s_t^{\mathrm{mkt}}$ includes open, high, low, close prices, trading volume, and basic technical indicators. Alignment is performed at the daily close level: a KOL signal published on day $t$ is paired with the market state observed at that day's close, and the corresponding action is evaluated from day $t{+}1$ onward. Non-trading days are forward-filled to the next trading session.

\vspace{0.05in}\noindent\textbf{Trajectory Construction}.
KOL-conditioned trading trajectories are constructed by pairing each signal-aligned baseline action $a_t^{\mathrm{base}}$ with the corresponding market state and subsequent price evolution. The baseline action is obtained by mapping the extracted sentiment and confidence signal into an executable anchor weight through the portfolio layer. Each trajectory consists of sequential transitions $(s_t, a_t, r_t, s_{t+1})$ where the reward $r_t$ reflects the realized portfolio return.
As described in Section~4, the resulting offline dataset exhibits strong silence domination: out of 1,229,021 total transitions, only 45,626 (3.71\%) carry an active KOL signal, while 1,183,395 (96.29\%) fall in the silence regime where no fresh KOL guidance is available.

\vspace{0.05in}\noindent\textbf{Benchmark Subset}.
For the experimental evaluation, we select a 20-KOL benchmark subset (10 per platform) based on signal density, asset coverage, and evaluation-period activity.
Table~\ref{tab:dataset_summary} summarizes the key dataset statistics.

\begin{figure}[!t]
    \centering
    \includegraphics[width=\columnwidth]{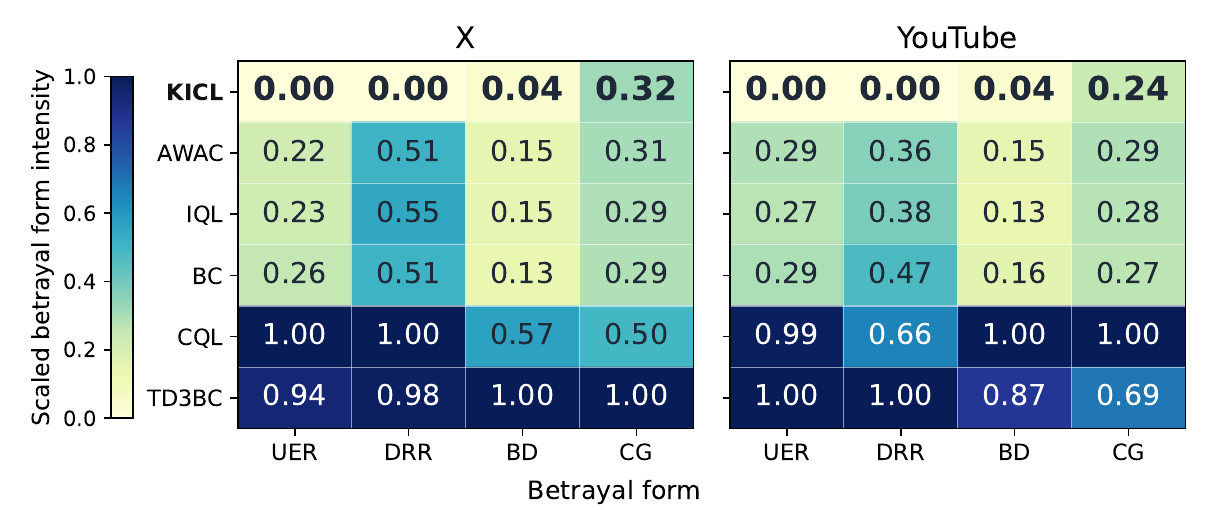}
    \caption{Scaled betrayal-form profiles on X and YouTube.}
    \label{fig:betrayal_form_profile}
\end{figure}

\subsection{Experiment Setup}

We compare KICL with three baseline groups: heuristics (RMB, HAP), imitation/supervised methods (BC \cite{bain1995framework}, SUP-DELTA), and generic offline RL methods (IQL \cite{kostrikov2021offline}, AWAC \cite{ashvin2020accelerating}, CQL \cite{kumar2020conservative}, and TD3+BC \cite{fujimoto2021minimalist}). All methods are trained and evaluated under an event-conditioned benchmark.

The model is exposed to only a compact set of six market factors---1-day and 5-day returns, 5-day and 20-day return volatility, a 20-day volume z-score, and distance to the 20-day moving average---together with KOL-derived signals. This keeps the setup intentionally conservative, so performance gains cannot be attributed mainly to richer market features. The KOL-aligned baseline is obtained by mapping sentiment and confidence scores into executable anchor weights through the portfolio layer, and serves as the intent anchor throughout replay construction, training, and evaluation. In benchmark tables, “Baseline” denotes this executable KOL-aligned anchor policy under the same decoding and evaluation pipeline. Replay behavior actions, by contrast, are the fitted dataset actions used for offline learning. The state combines text and ticker embeddings with core scalar features (sentiment, confidence, last position, and silence days) and the six market factors above.
We report both performance and intent-preservation metrics (Table~\ref{tab:eval_metrics}). Performance is evaluated by return, Sharpe ratio, and maximum drawdown, with win rate (WR) reported as an auxiliary event-level comparison against the executable KOL-aligned baseline. Intent preservation is evaluated by unsupported entry rate (UER), direction reversal rate (DRR), and baseline deviation (BD). We also report correlation gap (CG) as a diagnostic measure of same-direction co-movement mismatch relative to the baseline action sequence. Detailed configs are in the supplementary material.
\subsection{Results and Analysis}
Table~\ref{tab:main_results} reports the main results. On X, KICL delivers the best return and Sharpe ratio while maintaining zero unsupported entry and zero directional reversal. On YouTube, KICL achieves the best Sharpe ratio and win rate while preserving the same zero-hard-betrayal profile.
The grouped comparison reveals a clear pattern: generic offline RL baselines gain flexibility at the cost of substantially higher betrayal rates, while heuristic and imitation/supervised baselines preserve the KOL anchor but deliver smaller gains. KICL is the only method that combines top-tier return performance with zero unsupported entry and zero directional reversal on both platforms. This motivates a closer breakdown of how different methods diverge from the KOL-aligned baseline.

\vspace{0.05in}\noindent\textbf{Betrayal-Form Profile.}
We treat UER and DRR as \emph{hard betrayal}, BD as softer policy drift, and CG as same-direction co-movement mismatch. As shown in Figure~\ref{fig:betrayal_form_profile}, KICL drives UER and DRR to zero on both platforms while keeping BD low. Competing baselines gain flexibility through stronger hard betrayal.

\begin{figure}[!htbp]
    \centering
    \includegraphics[width=\columnwidth]{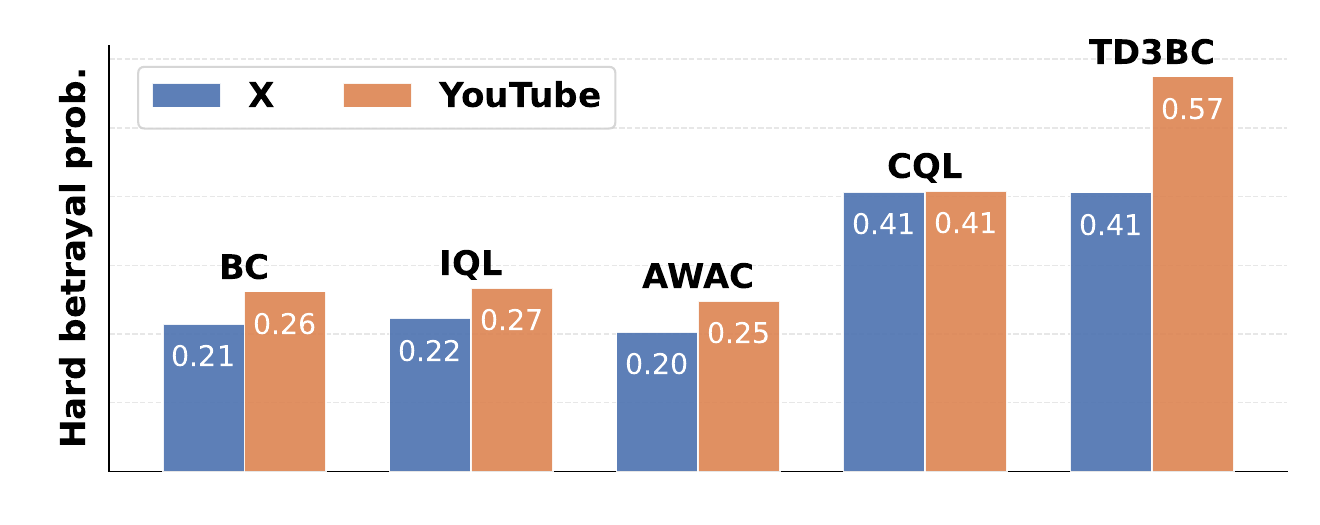}
    \caption{Hard-betrayal probability under excess-positive events on X and YouTube.}
    \label{fig:profit_linked_betrayal}
\end{figure}

\vspace{0.05in}\noindent\textbf{Profit-Linked Betrayal.}
Figure~\ref{fig:profit_linked_betrayal} reports hard-betrayal probability under excess-positive events. Hard betrayal is consistently nontrivial, especially for CQL and TD3+BC on YouTube; BC, IQL, AWAC, and HAP also show nonzero rates. KICL's gains arise from constrained execution refinement, not intent override.

\subsection{Ablation Study}

We organize the ablation in two parts. The first follows the actual construction path of the model, starting from the KOL-aligned baseline and adding the main structural components step by step until the full system is recovered. The second isolates where hard intent-preservation constraints must act: training, inference, or both.

\begin{table}[!htbp]
\small
\centering
\setlength{\tabcolsep}{3pt}
\caption{Component structure of the ablation variants.}
\label{tab:ablation_structure}
\begin{tabular}{l c c c c c}
\toprule
\textbf{Variant} & \textbf{Anchor} & \textbf{Policy} & \textbf{RL Comp.} & \textbf{Regime} & \textbf{Hard} \\
\midrule
Baseline & \checkmark &  &  &  & \checkmark \\
WO-RL-COMPLETION & \checkmark & \checkmark &  &  & \checkmark \\
WO-REGIME-SPLIT & \checkmark & \checkmark & \checkmark &  & \checkmark \\
WO-H & \checkmark & \checkmark & \checkmark & \checkmark &  \\
\textbf{KICL (Full)} & \textbf{\checkmark} & \textbf{\checkmark} & \textbf{\checkmark} & \textbf{\checkmark} & \textbf{\checkmark} \\
\bottomrule
\end{tabular}
\end{table}

Table~\ref{tab:ablation_structure} summarizes the structural path of the ablation. Starting from the executable KOL-aligned baseline, we first add a learnable policy shell, then introduce RL-based completion, and finally recover the full model by adding regime split. Hard constraints are retained throughout this constructive path and removed only in WO-H, which is treated separately as a destructive control rather than a normal intermediate variant.

\paragraph{Ablation I: Progressive completion path.}

\begin{table}[!htbp]
\small
\centering
\setlength{\tabcolsep}{3pt}
\caption{Event-level ablation results relative to the KOL-aligned baseline.}
\label{tab:ablation_main}
\begin{tabular}{l c c c c}
\toprule
\textbf{Variant} & $\Delta$\textbf{Return} & $\Delta$\textbf{Sharpe} & \textbf{MDD Change} & $\Delta$\textbf{BD} \\
\midrule
Baseline & -- & -- & -- & -- \\
WO-RL-COMPLETION & -12.2\% & -6.1\% & +0.3\% & +0.0001 \\
WO-REGIME-SPLIT  & +6.9\% & -4.2\% & +1.0\% & +0.0083 \\
WO-H             & -65.8\% & -29.8\% & +16.9\% & +0.0306 \\
\textbf{KICL (Full)} & \textbf{+18.9\%} & \textbf{+1.1\%} & +7.1\% & +0.0077 \\
\bottomrule
\end{tabular}
\end{table}

Table~\ref{tab:ablation_main} summarizes the event-level gains of the progressive ablation path relative to the KOL-aligned baseline.

\begin{figure}[!htbp]
    \centering
    \includegraphics[width=\columnwidth]{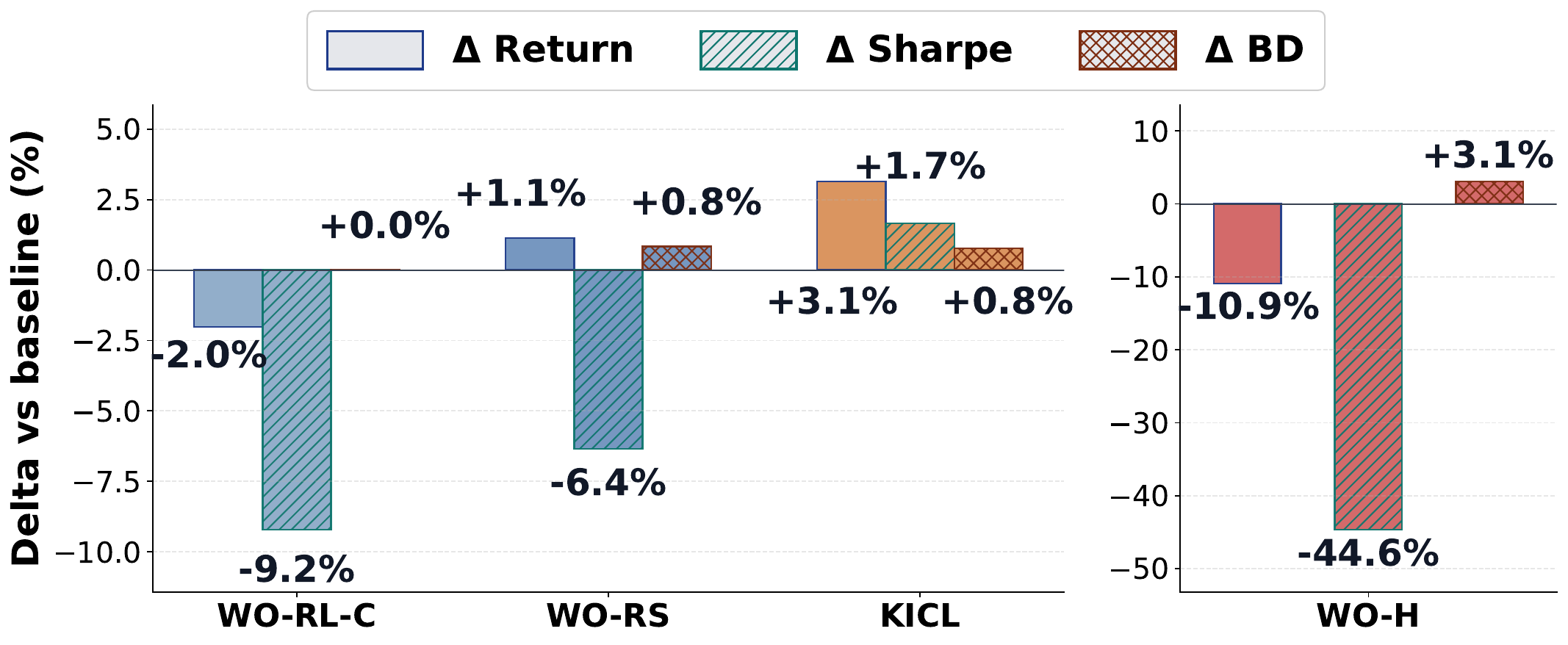}
    \caption{Progressive ablation relative to the KOL-aligned baseline. WO-H is shown separately as a destructive control. Bars denote $\Delta$Return, $\Delta$Sharpe, and $\Delta$BD.}
    \label{fig:ablation_performance}
\end{figure}

Figure~\ref{fig:ablation_performance} shows that the gain is not obtained in a single step, but through the staged construction of the full system. WO-RL-COMPLETION remains below the baseline in both return and Sharpe, indicating that a learnable policy shell alone is not sufficient. Adding RL-based completion recovers return, but the full model only emerges after regime split, which yields the strongest overall trade-off. WO-H is clearly disconnected from this constructive path: once hard barriers are removed, return and Sharpe collapse while baseline deviation rises sharply. This pattern supports the intended design logic of KICL as a progressively assembled completion system rather than a loosely combined set of components.

\paragraph{Ablation II: Where hard constraints must act.}

We compare four settings in which hard constraints are applied at training only, inference only, both, or neither.

\begin{figure}[!t]
    \centering
    \includegraphics[width=\columnwidth]{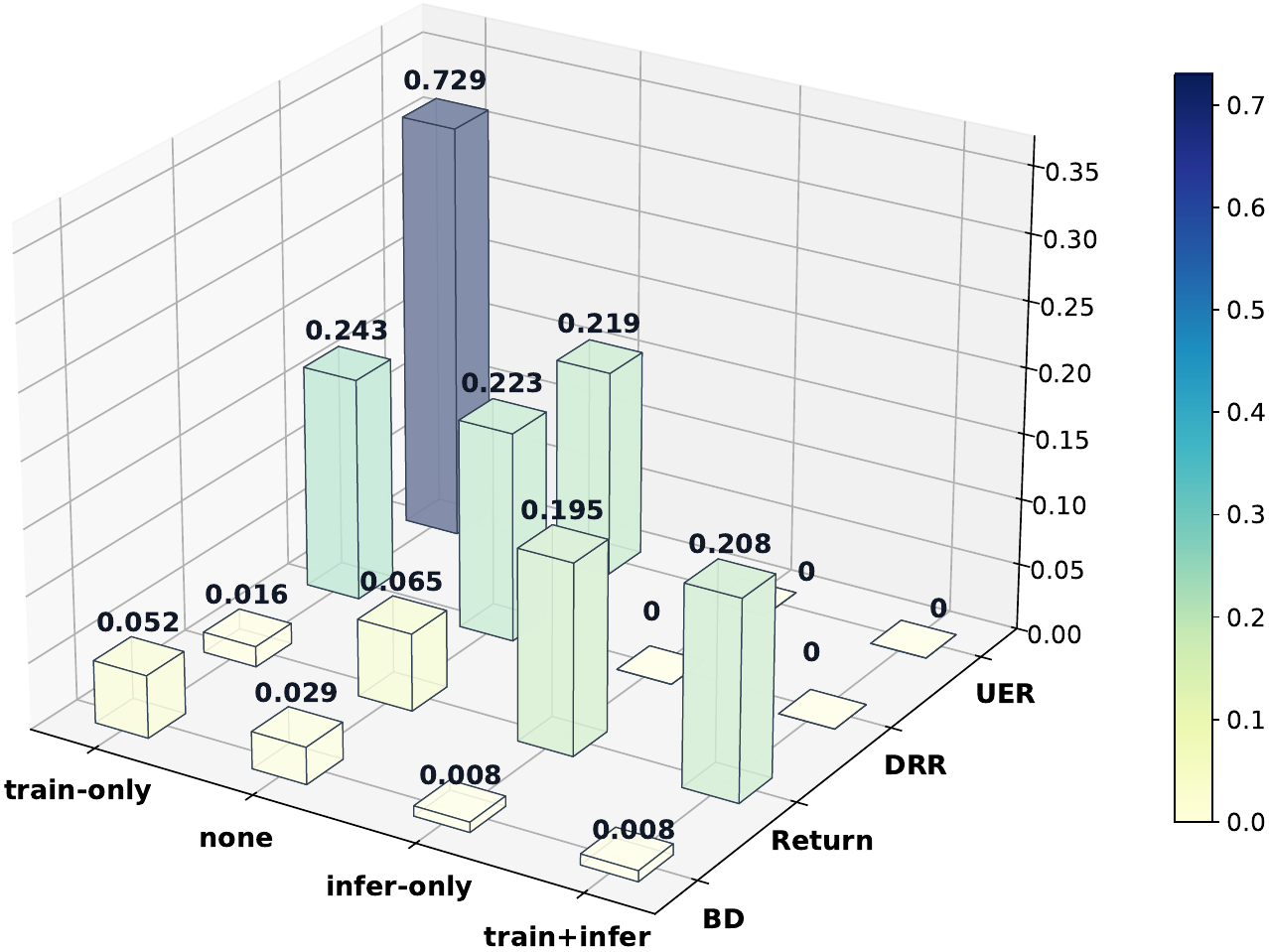}
    \caption{Hard-scope ablation across train-only, infer-only, train+infer, and no-hard settings.}
    \label{fig:hard_scope}
\end{figure}

Figure~\ref{fig:hard_scope} shows that the decisive effect comes from inference-time enforcement. The infer-only and train+infer settings stay close to each other across return and betrayal metrics, whereas train-only behaves much more like no-hard than like the constrained settings. In other words, soft regularization during training is not enough to preserve intent at action realization time; the boundary must be enforced when actions are actually decoded. Training-time constraints are therefore not entirely useless, but their contribution is secondary: they may help reduce softer drift, while the decisive protection against unsupported entry and reversal comes from inference-time enforcement.

\FloatBarrier

\subsection{Case Study}

\begin{figure}[!htbp]
    \centering
    \includegraphics[width=\columnwidth]{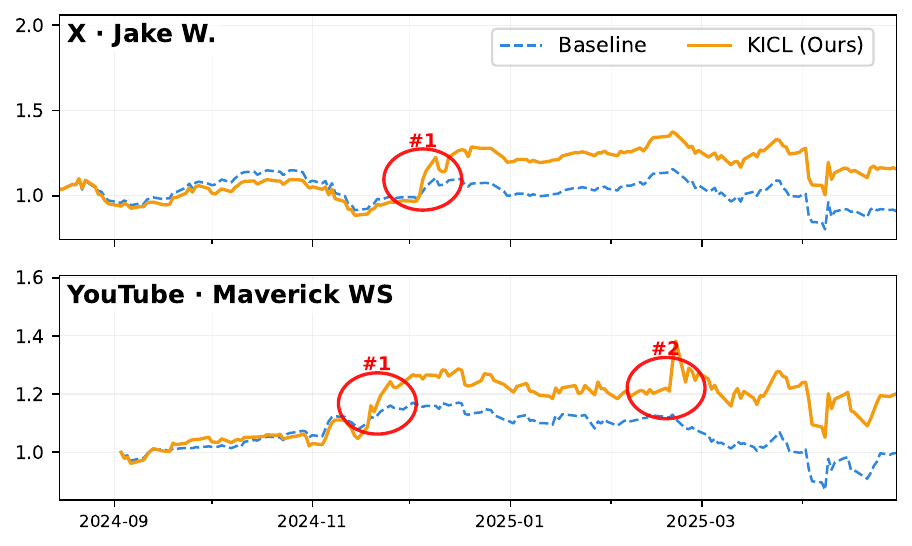}
    \caption{Representative case comparisons against the KOL-aligned baseline on X and YouTube; red circles mark the divergence nodes.}
    \label{fig:case_study}
\end{figure}

We revisit the same FXI and TSLA discourse contexts used in Figure~\ref{fig:kol_partial_policy_example}. These two examples illustrate two different forms of completion: amplifying exposure under a stable bullish cue, and restraining exposure under a neutral or wait-oriented cue. The red circles in Figure~\ref{fig:case_study} mark the divergence nodes where KICL starts to separate from the rule-based KOL baseline.

\begin{table}[!t]
\centering
\scriptsize
\setlength{\tabcolsep}{2.5pt}
\renewcommand{\arraystretch}{1.04}
\caption{Exposure-path evidence at the FXI and TSLA divergence nodes in Figure~\ref{fig:case_study}. Positive values denote long exposure and zero denotes near-flat allocation.}
\label{tab:case_fxitsla}
\begin{tabular*}{\columnwidth}{@{\extracolsep{\fill}} l c c c l @{}}
\toprule
\textbf{Case} & \textbf{Date} & \textbf{Base} & \textbf{Ours} & \textbf{Takeaway} \\
\midrule
\multicolumn{5}{@{}l@{}}{\textit{FXI (X/Jake, bullish cue)}} \\
FXI  & 2024-12-02 & 0.0554 & 0.0938 & Higher long exposure \\
FXI  & 2024-12-23 & 0.0000 & 0.0000 & Both close \\
FXI  & 2024-12-24 & 0.1321 & 0.1558 & Higher-weight re-entry \\
FXI  & 2024-12-26 & 0.0000 & 0.0000 & Both close again \\
\midrule
\multicolumn{5}{@{}l@{}}{\textit{TSLA (YT/Maverick, neutral/wait cue)}} \\
TSLA & 2024-11-11 & 0.0000 & 0.0000 & Both stay flat \\
TSLA & 2024-11-27 & 0.0554 & 0.0483 & Slightly more conservative \\
TSLA & 2024-12-13 & 0.0615 & 0.0281 & Much more conservative \\
TSLA & 2024-12-24 & 0.1795 & 0.1754 & Similar, still lower \\
TSLA & 2025-01-06 & 0.1817 & 0.1215 & Exposure remains capped \\
TSLA & 2025-01-21 & 0.0000 & 0.0000 & Both close \\
\bottomrule
\end{tabular*}
\end{table}

At the FXI node, KICL preserves the bullish anchor and increases long exposure (0.0938 vs.\ 0.0554 on 2024-12-02; 0.1558 vs.\ 0.1321 on 2024-12-24), showing that completion can strengthen execution when market context remains aligned with the original positive cue. At the TSLA node, KICL follows a more conservative exposure path under the same neutral/wait stance (0.0281 vs.\ 0.0615 on 2024-12-13; 0.1215 vs.\ 0.1817 on 2025-01-06), showing that completion can also restrain exposure without introducing an opposite directional trade. Taken together, the two cases indicate that KICL adjusts execution intensity in both directions---amplification and restraint---while preserving the semantic anchor of the discourse.

\section{Discussion and Future Work}

The main results and ablations point to the same conclusion: financial KOL discourse is useful as directional guidance, but executable performance depends on completing the missing execution structure under explicit intent constraints. KICL’s gains come from constrained completion of missing execution structure, not unrestricted policy generation: it improves returns while maintaining zero unsupported entry and zero directional reversal. The ablations reinforce this—performance improves along the constrained path, while removing hard barriers sharply degrades results, indicating that intent preservation is a structural requirement rather than a soft preference in this setting.

This study is intentionally conservative: the market state uses only basic features, and the completion mechanism adopts a simple baseline-plus-residual design. This makes the results easier to attribute to intent-preserving completion itself rather than to heavy market modeling. Future work could incorporate richer market context, stronger semantic grounding, and more expressive architectures, and test whether intent-preserving completion extends to other forms of partially specified language guidance.

\section{Conclusion}

This paper studies how to turn financial KOL discourse into executable trading policies without overriding the original intent. We show that such discourse is better understood as a partial trading policy and should be learned through intent-preserving completion around a KOL-aligned baseline. This perspective is largely absent from prior work because social-media-based studies typically treat language as an exogenous signal, while RL-based trading methods learn policies directly from market states. Our framework connects these two lines by completing partially specified policy guidance under explicit intent-preservation constraints. Empirically, KICL achieves strong event-level performance while maintaining the most consistent intent-preservation profile; the ablations confirm that its gains come from constrained execution refinement rather than replacement of the semantic anchor.

\bibliographystyle{ACM-Reference-Format}
\bibliography{refs}

\clearpage

\onecolumn
\appendix

\hypersetup{pageanchor=false}

\begin{titlepage}
\thispagestyle{empty}
\centering
\vspace*{\fill}
{\LARGE Supplementary Material\par}
\vspace{1.2em}
{\large This document contains supplementary materials for the main manuscript.\par}
\vspace*{\fill}
\end{titlepage}

\clearpage
\renewcommand{\thepage}{S\arabic{page}}
\setcounter{page}{1}
\setcounter{section}{0}
\setcounter{table}{0}
\setcounter{figure}{0}
\setcounter{equation}{0}
\renewcommand{\thesection}{S\arabic{section}}
\renewcommand{\thetable}{S\arabic{table}}
\renewcommand{\thefigure}{S\arabic{figure}}
\renewcommand{\theequation}{S\arabic{equation}}
\renewcommand{\theHsection}{supp.\arabic{section}}
\renewcommand{\theHsubsection}{supp.\arabic{section}.\arabic{subsection}}
\renewcommand{\theHtable}{supp.\arabic{table}}
\renewcommand{\theHfigure}{supp.\arabic{figure}}
\renewcommand{\theHequation}{supp.\arabic{equation}}
\hypersetup{pageanchor=true}

\vspace{1em}

\section{Full-Dataset Statistics of KOL Discourse}
\label{app:full_kol_stats}

To complement the source-level evidence reported in the main text, we provide a more comprehensive set of full-dataset statistics here. Unlike the selected benchmark subset used for training and evaluation, this full-dataset view is intended to characterize the broader empirical structure of financial KOL discourse itself. The results further support the same conclusion: KOL discourse is highly asset-specific and directional, yet execution-complete descriptions remain extremely rare, reinforcing the partial-policy interpretation.

\begin{center}
\small
\setlength{\tabcolsep}{4pt}
\begin{tabular}{lcc}
\toprule
\textbf{Statistic} & \textbf{YouTube} & \textbf{X} \\
\midrule
\multicolumn{3}{l}{\textit{Coverage / specificity}} \\
Mentioned companies & 6,774 & 3,811 \\
Ticker-explicit ratio & 81.61\% & 79.51\% \\
Directional-expression ratio & 79.58\% & 91.62\% \\
Conditional-action ratio & 47.19\% & 19.01\% \\
Single-mention ratio & 62.06\% & 39.67\% \\
\midrule
\multicolumn{3}{l}{\textit{Execution under-specification}} \\
Explicit entry ratio & 33.14\% & 5.48\% \\
Explicit sizing ratio & 12.48\% & 0.62\% \\
Explicit holding-horizon ratio & 13.63\% & 3.69\% \\
Explicit exit ratio & 14.59\% & 2.08\% \\
Full-execution completeness ratio & 1.86\% & 0.00\% \\
\midrule
\multicolumn{3}{l}{\textit{Temporal maintenance}} \\
No follow-up within 30 days & 89.58\% & 74.65\% \\
No follow-up within 60 days & 85.75\% & 69.82\% \\
No follow-up within 90 days & 82.88\% & 65.81\% \\
Silence duration (p90) & 107.13 days & 55.00 days \\
Sentiment reversal rate & 58.21\% & 49.69\% \\
Median time to first reversal & 179.84 days & 161.00 days \\
\midrule
\multicolumn{3}{l}{\textit{Sentiment distribution}} \\
Positive ratio & 49.84\% & 62.47\% \\
Negative ratio & 31.50\% & 29.74\% \\
Neutral ratio & 18.66\% & 7.78\% \\
\bottomrule
\end{tabular}

\captionof{table}{Full-dataset empirical characteristics of financial KOL discourse at the company level on YouTube and X (18 YouTube KOLs and 14 X KOLs).}
\label{tab:kol_discourse_stats_full}
\end{center}
\newpage

\newpage

\section{Additional Experimental Details}
\label{app:metric_defs}

\subsection{Detailed Definitions of Evaluation Metrics}

This appendix provides the exact definitions of the evaluation metrics reported in the main text. We divide them into conventional trading metrics and betrayal metrics relative to the KOL-aligned baseline action.

\paragraph{Conventional trading metrics.}
Let $r_t$ denote the daily portfolio return at time $t$, and let the evaluation horizon contain $T$ trading days.

\emph{Cumulative Return.} The cumulative return over the full evaluation period is defined as
\begin{equation}
\mathrm{CR} = \prod_{t=1}^{T}(1+r_t)-1.
\end{equation}
\vspace{0.5em}

\emph{Sharpe Ratio.} The annualized Sharpe ratio is computed as
\begin{equation}
\mathrm{Sharpe}=
\begin{cases}
\frac{\mu_r}{\sigma_r}\sqrt{252}, & \sigma_r>0,\\[4pt]
0, & \sigma_r=0,
\end{cases}
\end{equation}
where $\mu_r=\frac{1}{T}\sum_{t=1}^{T} r_t$ is the mean daily return and $\sigma_r$ is the standard deviation of daily returns.
\vspace{0.5em}

\emph{Maximum Drawdown.} Define the cumulative equity curve as
\begin{equation}
E_t=\prod_{i=1}^{t}(1+r_i),
\end{equation}
and the running peak as
\begin{equation}
P_t=\max_{1\le i\le t} E_i.
\end{equation}
Maximum drawdown is then given by
\begin{equation}
\mathrm{MDD}=\max_{1\le t\le T}\frac{P_t-E_t}{P_t+10^{-8}}.
\end{equation}

\vspace{0.8em}

\paragraph{Betrayal metrics.}
Let $a_t^{\mathrm{base}}$ denote the KOL-aligned baseline action and $a_t^\pi$ denote the learned policy action. We use two thresholds: $\tau_{\mathrm{entry}}$ for determining whether the baseline contains an active KOL signal, and $\tau_{\mathrm{act}}$ for determining whether the learned policy takes a nontrivial action. We define
\begin{equation}
\mathrm{HasSignal}_t = \mathbf{1}\!\left[|a_t^{\mathrm{base}}|\ge \tau_{\mathrm{entry}}\right],
\qquad
\mathrm{NoSignal}_t = \mathbf{1}\!\left[|a_t^{\mathrm{base}}|< \tau_{\mathrm{entry}}\right].
\end{equation}
\vspace{0.5em}

\emph{Unsupported Entry Rate (UER).} UER measures how often the learned policy opens a nontrivial position when no active KOL baseline signal is present:
\begin{equation}
\mathrm{UER}
=
\frac{
\sum_{t=1}^{T}
\mathbf{1}\!\left[
|a_t^{\mathrm{base}}|<\tau_{\mathrm{entry}}
\ \land\
|a_t^\pi|>\tau_{\mathrm{act}}
\right]
}{
\sum_{t=1}^{T}
\mathbf{1}\!\left[
|a_t^{\mathrm{base}}|<\tau_{\mathrm{entry}}
\right]
+10^{-8}
}.
\end{equation}
\vspace{0.5em}

\emph{Direction Reversal Rate (DRR).} DRR measures how often the learned policy takes the opposite direction to the baseline when an active KOL signal exists:
\begin{equation}
\mathrm{DRR}
=
\frac{
\sum_{t=1}^{T}
\mathbf{1}\!\left[
a_t^{\mathrm{base}} a_t^\pi < 0
\ \land\
|a_t^{\mathrm{base}}|\ge \tau_{\mathrm{entry}}
\right]
}{
\sum_{t=1}^{T}
\mathbf{1}\!\left[
|a_t^{\mathrm{base}}|\ge \tau_{\mathrm{entry}}
\right]
+10^{-8}
}.
\end{equation}
\vspace{0.5em}

\emph{Baseline Deviation (BD).} BD measures the average absolute deviation between the learned policy action and the baseline action:
\begin{equation}
\mathrm{BD}
=
\frac{1}{T}\sum_{t=1}^{T}\left|a_t^\pi-a_t^{\mathrm{base}}\right|.
\end{equation}

\vspace{0.5em}

\paragraph{Interpretation.}
For the conventional trading metrics, higher cumulative return and Sharpe ratio are preferred, while lower MDD is preferred. For the betrayal metrics, lower UER, DRR, and BD indicate stronger preservation of KOL intent.

\subsection{Training and Evaluation Configuration}
\label{app:training_config}

This subsection summarizes the concrete training and testing setup used by the current codebase. Unless otherwise noted, the values below are default settings in training/evaluation scripts and can be overridden by launcher flags for specific experiment groups.

\paragraph{Experiment scope and data sources.}
\begin{itemize}
\item Two datasets are used: a selected-20 benchmark subset and a full multisource dataset.
\item The same training and evaluation pipeline is applied consistently across methods.
\end{itemize}

\paragraph{Chronological split protocol.}
Data are split chronologically by \texttt{trading\_day}, with same-day samples kept in the same split. The ratio is train/val/test $=0.6/0.2/0.2$.

\paragraph{Trading-day alignment and no-leakage rule.}
All discourse events are first mapped to a unified market cutoff day before factor construction and replay generation. Posts published before market open or during trading hours are aligned to the same trading day, while after-hours and non-trading-day posts are aligned to the next trading session. Market factors are computed strictly from historical data up to the sample cutoff day. Concretely, if $P_t$ denotes the close price on the aligned cutoff day $t$, then all factor computation uses only the historical slice up to $t$, with no future observations beyond the sample day. Rows are dropped when historical warmup is insufficient for valid factor construction.

\paragraph{Market information exposed to the model.}
The model is exposed to a compact 6-dimensional market factor set rather than a large raw price window. Let $P_t$ denote the close price on day $t$, $V_t$ denote trading volume, and let the daily return be
\begin{equation}
r_t=\frac{P_t}{P_{t-1}}-1.
\end{equation}
The six market factors are defined as follows:
\begin{align}
\texttt{ret\_1d} &= \frac{P_t}{P_{t-1}}-1,\\
\texttt{ret\_5d} &= \frac{P_t}{P_{t-5}}-1,\\
\texttt{vol\_5d} &= \mathrm{Std}\!\left(r_{t-4},\dots,r_t\right),\\
\texttt{vol\_20d} &= \mathrm{Std}\!\left(r_{t-19},\dots,r_t\right),\\
\texttt{volu\_z\_20d} &= \frac{V_t-\mu_V^{(20)}(t)}{\sigma_V^{(20)}(t)},\\
\texttt{dist\_sma20} &= \frac{P_t}{\mathrm{SMA}_{20}(t)}-1,
\end{align}
where
\begin{equation}
\mu_V^{(20)}(t)=\frac{1}{20}\sum_{i=t-19}^{t}V_i,
\qquad
\sigma_V^{(20)}(t)=\mathrm{Std}\!\left(V_{t-19},\dots,V_t\right),
\end{equation}
and
\begin{equation}
\mathrm{SMA}_{20}(t)=\frac{1}{20}\sum_{i=t-19}^{t}P_i.
\end{equation}
In the current implementation, volatility uses population standard deviation ($\mathrm{ddof}=0$), and $\texttt{volu\_z\_20d}$ falls back to $0.0$ when the volume standard deviation is numerically zero.

\paragraph{State representation.}
Each training state is constructed as the concatenation of text embedding, ticker embedding, core execution-context scalars, and compact market factors:
\begin{equation}
s_t=
\bigl[
e_t^{\text{text}}
\ \Vert\
e_t^{\text{ticker}}
\ \Vert\
x_t^{\text{core}}
\ \Vert\
x_t^{\text{mkt}}
\bigr],
\end{equation}
where the core scalar features are
\begin{equation}
x_t^{\text{core}}
=
[\texttt{sentiment},\ \texttt{confidence},\ \texttt{last\_position},\ \texttt{silence\_days}],
\end{equation}
and the market factor vector is
\begin{equation}
x_t^{\text{mkt}}
=
[\texttt{ret\_1d},\ \texttt{ret\_5d},\ \texttt{vol\_5d},\ \texttt{vol\_20d},\ \texttt{volu\_z\_20d},\ \texttt{dist\_sma20}].
\end{equation}
Thus, market information enters the model only through this compact factor set rather than through an unrestricted raw price sequence.

\paragraph{Baseline action construction and evaluation.}
The KOL-aligned baseline action is built from the extracted sentiment and confidence signal. For each discourse row, a bounded raw baseline score is first computed as
\begin{equation}
a^{\text{raw}}_t = \tanh\!\bigl(2 \cdot \texttt{sentiment}_t \cdot \texttt{confidence}_t\bigr).
\end{equation}
This raw score is then mapped into an executable portfolio-level baseline action through the portfolio layer, which normalizes exposure, applies per-asset caps, and preserves continuity through previous portfolio state. The resulting \texttt{baseline\_weight} is used as the replay-level intent anchor and stored as \texttt{baseline\_actions} in the replay buffer.

The replay buffer also stores a separate behavior action, denoted \texttt{actions}, which is obtained from a smoothed execution path built around the baseline signal. In this sense, the baseline action serves as the semantic intent anchor, while the behavior action serves as the offline action target for BC/IQL fitting.

During policy learning and inference, the action is parameterized residually around the baseline:
\begin{equation}
a_t^\pi = a_t^{\mathrm{base}} + \delta_t,
\end{equation}
with hard intent constraints optionally applied to forbid unsupported entry when $|a_t^{\mathrm{base}}|$ is below threshold and to forbid direction reversal against the baseline sign.

For baseline evaluation in benchmark comparison, the same decoding and constraint pipeline is used with a zero residual actor ($\delta_t=0$). Therefore, ``Baseline'' in the benchmark tables refers to the executable KOL-aligned anchor policy under the same portfolio and evaluation mechanics, rather than an unrelated external heuristic.

\paragraph{KICL architecture.}
KICL uses a residual-action design:
\begin{itemize}
\item \textbf{Actor (dual-head, deterministic):} backbone MLP \,$state\_dim\!\to\!512\to\!512\to\!256$ (ReLU), with two \texttt{Tanh} heads for \texttt{delta\_signal} and \texttt{delta\_decay}.
\item \textbf{Critic (Q):} input concatenates state, baseline action, and delta action (effective dimension \,$state\_dim+2$); hidden layers are \,$512\to512\to256$, output dimension 1.
\item \textbf{Value:} input concatenates state and baseline action (effective dimension \,$state\_dim+1$); hidden layers are \,$512\to512\to256$, output dimension 1.
\end{itemize}

\paragraph{Default training hyperparameters (KICL).}
{\small
\begin{center}
\setlength{\tabcolsep}{3.5pt}
\renewcommand{\arraystretch}{0.93}
\begin{tabular}{p{2.7cm}p{7.4cm}p{2.3cm}}
\toprule
\textbf{Group} & \textbf{Hyperparameter} & \textbf{Default} \\
\midrule
BC pretraining & epochs & 10 \\
BC pretraining & batch size & 256 \\
BC pretraining & learning rate & $3\times10^{-4}$ \\
BC pretraining & behavior fitting enabled & True \\
BC pretraining & anchor loss weight & 0.03 \\
\midrule
IQL optimization & training steps & 100000 \\
IQL optimization & batch size & 256 \\
IQL optimization & actor / critic / value learning rates & $3\times10^{-4}$ / $3\times10^{-4}$ / $3\times10^{-4}$ \\
IQL optimization & discount factor $\gamma$ & 0.99 \\
IQL optimization & expectile $\tau$ & 0.7 \\
IQL optimization & temperature coefficient $\beta$ & 3.0 \\
\midrule
Intent constraints & fidelity loss weight & 0.03 \\
Intent constraints & actor alignment weight & 0.04 \\
Intent constraints & unsupported-entry penalty weight & 0.02 \\
Intent constraints & reversal penalty weight & 0.05 \\
Intent constraints & entry threshold & $5\times10^{-4}$ \\
Intent constraints & delta clamp range & 1.8 \\
Intent constraints & hard intent constraints enabled & True \\
Intent constraints & regime split enabled & True \\
\midrule
Market/runtime & market factors enabled & True \\
Market/runtime & market factor dimension & 6 \\
Market/runtime & logging interval & 200 \\
Market/runtime & write IQL CSV log & True \\
Market/runtime & device selection & CUDA if available \\
\bottomrule
\end{tabular}
\end{center}
}

\paragraph{Ablation and benchmark overrides.}\vspace{-0.6em}
{\footnotesize
\begin{itemize}
\setlength{\itemsep}{0.1em}
\setlength{\parskip}{0pt}
\setlength{\parsep}{0pt}
\setlength{\topsep}{0.15em}
\setlength{\partopsep}{0pt}
\item KICL ablations remove hard intent constraints, soft penalties, BC-anchor regularization, RL completion, selected penalties, market factors, or regime split.
\item Hard-scope settings are \texttt{hard\_both}, \texttt{hard\_train\_only}, \texttt{hard\_infer\_only}, and \texttt{hard\_none}; typical values are BC epochs $=10$, IQL steps $=100000$, and batch size $=256$.
\item Baselines include BC, IQL, RMB, HAP, SUP\_DELTA, CQL, AWAC, and TD3BC. Typical settings use batch size $=256$ and steps $=100000$; CQL uses $\alpha=1.0$, temperature $=1.0$, sampled actions $=10$; AWAC uses $\beta=1.0$ and max weight $=20.0$.
\end{itemize}
}

\clearpage

\section{Unified Financial KOL Selection Standard}

To ensure that the KOL pool used in this work is not manually tailored to the paper, we adopted a unified selection standard designed to be public, general, transparent, and reproducible. The goal of this standard is not to optimize for the downstream benchmark outcome, but to provide a stable and platform-consistent rule for identifying financially meaningful KOLs. The same standard was used to construct the candidate pools from which the final X and YouTube benchmark KOL sets were selected.

The standard consists of three components: finance-topic discovery, network validation, and quantitative ranking. Together, these steps are intended to ensure that selected KOLs are not only active in finance-related discourse, but also exhibit sustained audience response and structural embeddedness in the broader financial influencer community.

\subsection{Stage 1: Finance-Topic Discovery}

We first identify candidate KOLs through finance-topic discovery. The purpose of this stage is to ensure that candidate authors are genuinely active in financial discourse rather than being generic high-visibility accounts with occasional market-related content.

For X, candidate posts are collected through finance-indicative search queries over a recent time window. The query set covers representative categories of market discourse, including equity cashtags, crypto-related terms, macroeconomic expressions, earnings and valuation language, and market-structure terminology. To suppress low-information noise, only posts satisfying minimum engagement thresholds are retained:
\begin{equation}
\texttt{replyCount} \ge 20,
\qquad
\texttt{likeCount} \ge 200.
\end{equation}

Returned posts are then aggregated at the author level. For each author $u$, we compute summary statistics including the number of posts, the number of active days, and median views. An author is retained as an initial candidate only if
\begin{equation}
n_{\text{posts}}(u) \ge 3,
\qquad
n_{\text{active days}}(u) \ge 3,
\qquad
\text{median\_views}(u) \ge 3000.
\end{equation}

This stage yields an initial set of finance-relevant and sufficiently active candidate authors. While the concrete data source differs across platforms, the same principle is applied to both X and YouTube: authors must show sustained, finance-centered activity with nontrivial audience response, rather than isolated or weakly related mentions.

\subsection{Stage 2: Network Validation}

Content relevance alone is insufficient for KOL selection, since keyword-based discovery may still include noisy or peripheral accounts. We therefore add a network-validation stage to test whether a candidate author is structurally embedded in the financial influencer community.

Starting from the initial seed set, we examine local follower-network relations and identify candidate accounts that either follow multiple seed authors or are followed by multiple seed authors. Let $S$ denote the current seed-author set, and let $m(u)$ denote the number of mutual seed overlaps associated with author $u$. We retain only candidates satisfying
\begin{equation}
m(u) \ge 5.
\end{equation}

This rule serves as a structural consistency filter: a candidate is not selected merely because of finance-related wording, but also because it occupies a nontrivial position within the broader finance-discourse network.

\subsection{Stage 3: Quantitative Ranking}

After topic discovery and network validation, all remaining candidates are ranked by a quantitative score that combines semantic finance relevance with engagement quality. The purpose of this stage is to provide a stable and transparent ordering rule for candidate KOLs.

\paragraph{Finance relevance.}
For each author, recent posts are vectorized and compared against a predefined finance keyword set. Let $\mathrm{post}_i$ denote the $i$-th post of an author, and let $\mathrm{keyphrase}_k$ denote the $k$-th finance keyphrase. The raw semantic relevance score is defined as
\begin{equation}
\mathrm{sim}_{\mathrm{raw}}
=
\frac{1}{N}
\sum_{i=1}^{N}
\max_k
\Bigl(
\cos(\mathrm{post}_i,\mathrm{keyphrase}_k)
\Bigr),
\end{equation}
where $N$ is the number of posts used for scoring. Higher values indicate stronger semantic alignment with finance-related discourse.

\paragraph{Engagement quality.}
For each post, we define the engagement score as
\begin{equation}
ER_i
=
\texttt{likes}_i
+
\texttt{replies}_i
+
\texttt{retweets}_i
+
\texttt{quotes}_i.
\end{equation}
We then average over the author's recent posts:
\begin{equation}
ER_{\mathrm{avg}}
=
\frac{1}{N}
\sum_{i=1}^{N} ER_i.
\end{equation}
To reduce the dominance of extremely large accounts, we apply logarithmic compression:
\begin{equation}
\mathrm{eng}_{\mathrm{raw}}
=
\log\!\bigl(1 + ER_{\mathrm{avg}} \cdot 1000\bigr).
\end{equation}

\paragraph{Normalization and final score.}
Both scores are min--max normalized across eligible candidates:
\begin{equation}
\mathrm{sim}_{\mathrm{norm}}
=
\frac{\mathrm{sim}_{\mathrm{raw}}-\min(\mathrm{sim})}
{\max(\mathrm{sim})-\min(\mathrm{sim})},
\qquad
\mathrm{eng}_{\mathrm{norm}}
=
\frac{\mathrm{eng}_{\mathrm{raw}}-\min(\mathrm{eng})}
{\max(\mathrm{eng})-\min(\mathrm{eng})},
\end{equation}
so that
\begin{equation}
\mathrm{sim}_{\mathrm{norm}},\ \mathrm{eng}_{\mathrm{norm}} \in [0,1].
\end{equation}

The final ranking score is defined as
\begin{equation}
\mathrm{score}
=
0.4 \cdot \mathrm{sim}_{\mathrm{norm}}
+
0.6 \cdot \mathrm{eng}_{\mathrm{norm}}.
\end{equation}
This weighting gives slightly greater emphasis to sustained audience response while preserving finance-topic relevance as an explicit component of the selection standard.

\subsection{Eligibility and Standardized Selection Rule}

To maintain consistency, candidate accounts are subject to the same eligibility filter before final ranking. In the handle-based ranking pipeline, accounts with larger established audience reach are prioritized through a follower-based eligibility rule:
\begin{equation}
\texttt{followers} > 200{,}000.
\end{equation}
Accounts below this threshold are retained in the broader pool but assigned lower ranking priority.

The resulting selection standard is therefore not based on any single attribute such as popularity, follower count, or keyword frequency. Instead, a KOL must satisfy all three dimensions of the standard:
\begin{enumerate}
    \item explicit financial-topic relevance,
    \item nontrivial and sustained audience engagement,
    \item structural validation within the financial influencer network.
\end{enumerate}

\subsection{Role in This Work}

We emphasize that this standard is used as a general and platform-consistent KOL selection rule rather than a paper-specific tuning device. Its purpose is to make the benchmark construction process public, transparent, and reproducible. Under this standard, the X-side and YouTube-side KOL lists are selected according to the same high-level criteria of finance relevance, sustained activity, engagement quality, and community embeddedness. In this sense, the standard defines the admissible KOL pool before any downstream model comparison is conducted, helping separate benchmark construction from result-driven manual selection.

\clearpage

\section{KOL-Level Benchmark Results}
\label{app:kol_level_results}

For consistency with the main benchmark table, we report the same event-level performance and betrayal metrics on X and YouTube. Within each KOL block, the best value in each column is shown in bold and the second-best value is underlined. Rows marked with $^{\star}$ denote our proposed method.

\begin{center}
\centering
\captionof{table}{KOL-level results on X.}
\label{tab:appendix_x_kol_results}
\footnotesize
\setlength{\tabcolsep}{1.6pt}
\renewcommand{\arraystretch}{1.80}
\begin{minipage}[t]{0.49\textwidth}
\centering
\resizebox{0.982\linewidth}{!}{
\begin{tabular}{>{\centering\arraybackslash}p{3.1cm} >{\centering\arraybackslash}p{1.35cm} cccccc}
\toprule
\textbf{KOL} & \textbf{Mth.} & \textbf{Ret.}$\uparrow$ & \textbf{Sha.}$\uparrow$ & \textbf{MDD}$\downarrow$ & \textbf{UER}$\downarrow$ & \textbf{DRR}$\downarrow$ & \textbf{BD}$\downarrow$ \\
\midrule
bespokeinvest & BC & -0.023 & -0.257 & \underline{0.060} & 0.028 & \underline{0.258} & 0.005 \\
 & IQL & -0.051 & -0.510 & 0.064 & \underline{0.023} & 0.368 & \underline{0.004} \\
 & CQL & \textbf{0.011} & \textbf{0.216} & \textbf{0.040} & 0.998 & 0.926 & 0.172 \\
 & TD3BC & \underline{-0.004} & \underline{-0.010} & 0.060 & 0.689 & 0.302 & 0.111 \\
 & AWAC & -0.047 & -0.526 & 0.060 & 0.024 & 0.340 & 0.004 \\
 & Ours$^{\star}$ & -0.111 & -0.223 & 0.291 & \textbf{0.000} & \textbf{0.000} & \textbf{0.001} \\
\midrule
EliteOptions2 & BC & 0.149 & 0.785 & 0.109 & 0.154 & 0.121 & 0.026 \\
 & IQL & 0.107 & 0.623 & 0.108 & 0.142 & 0.083 & 0.026 \\
 & CQL & \underline{0.162} & \underline{0.848} & \underline{0.106} & \underline{0.136} & 0.086 & 0.025 \\
 & TD3BC & -0.040 & -1.243 & \textbf{0.055} & 0.987 & 0.710 & 0.236 \\
 & AWAC & 0.121 & 0.672 & 0.108 & 0.139 & \underline{0.073} & \underline{0.024} \\
 & Ours$^{\star}$ & \textbf{0.256} & \textbf{0.899} & 0.184 & \textbf{0.000} & \textbf{0.000} & \textbf{0.013} \\
\midrule
goldseek & BC & \underline{-0.052} & -0.406 & 0.086 & 0.260 & 0.049 & 0.033 \\
 & IQL & -0.053 & \underline{-0.401} & 0.083 & 0.214 & 0.064 & 0.033 \\
 & CQL & -0.075 & -0.851 & \underline{0.068} & 0.950 & \underline{0.028} & 0.070 \\
 & TD3BC & -0.093 & -1.392 & \textbf{0.047} & 0.923 & 0.183 & 0.277 \\
 & AWAC & -0.059 & -0.468 & 0.085 & \underline{0.213} & 0.058 & \underline{0.032} \\
 & Ours$^{\star}$ & \textbf{0.625} & \textbf{1.851} & 0.276 & \textbf{0.000} & \textbf{0.000} & \textbf{0.008} \\
\midrule
GoldTelegraph & BC & -0.042 & -0.276 & 0.184 & 0.286 & \underline{0.019} & 0.034 \\
 & IQL & \underline{-0.033} & -0.195 & 0.182 & 0.232 & \underline{0.019} & 0.033 \\
 & CQL & -0.082 & -0.815 & \underline{0.171} & 0.973 & \textbf{0.000} & 0.104 \\
 & TD3BC & -0.109 & -1.401 & \textbf{0.128} & 0.989 & 0.304 & 0.287 \\
 & AWAC & -0.033 & \underline{-0.189} & 0.186 & \underline{0.223} & \underline{0.019} & \underline{0.032} \\
 & Ours$^{\star}$ & \textbf{0.050} & \textbf{0.434} & 0.374 & \textbf{0.000} & \textbf{0.000} & \textbf{0.005} \\
\midrule
intocryptoverse & BC & -0.068 & -0.682 & 0.160 & 1.000 & \textbf{0.000} & \underline{0.080} \\
 & IQL & -0.049 & -0.544 & \underline{0.131} & 0.864 & \textbf{0.000} & 0.110 \\
 & CQL & -0.052 & \underline{-0.220} & 0.229 & \underline{0.833} & \underline{0.007} & 0.118 \\
 & TD3BC & \textbf{0.057} & \textbf{0.719} & 0.158 & 0.862 & 0.029 & 0.156 \\
 & AWAC & \underline{-0.032} & -0.318 & \textbf{0.125} & 0.843 & \textbf{0.000} & 0.109 \\
 & Ours$^{\star}$ & -0.144 & -0.250 & 0.399 & \textbf{0.000} & \textbf{0.000} & \textbf{0.023} \\
\bottomrule
\end{tabular}
}
\end{minipage}\hfill
\begin{minipage}[t]{0.49\textwidth}
\centering
\resizebox{0.982\linewidth}{!}{
\begin{tabular}{>{\centering\arraybackslash}p{3.1cm} >{\centering\arraybackslash}p{1.35cm} cccccc}
\toprule
\textbf{KOL} & \textbf{Mth.} & \textbf{Ret.}$\uparrow$ & \textbf{Sha.}$\uparrow$ & \textbf{MDD}$\downarrow$ & \textbf{UER}$\downarrow$ & \textbf{DRR}$\downarrow$ & \textbf{BD}$\downarrow$ \\
\midrule
Jake\_Wujastyk & BC & 0.138 & 1.226 & 0.103 & 0.086 & 0.426 & 0.013 \\
 & IQL & 0.194 & 1.488 & 0.104 & \underline{0.084} & 0.360 & \underline{0.011} \\
 & CQL & 0.097 & 1.730 & \underline{0.031} & 0.706 & 0.213 & 0.058 \\
 & TD3BC & 0.104 & \underline{1.816} & \textbf{0.023} & 0.974 & \underline{0.005} & 0.245 \\
 & AWAC & \underline{0.287} & \textbf{1.883} & 0.103 & 0.085 & 0.314 & 0.011 \\
 & Ours$^{\star}$ & \textbf{0.462} & 1.399 & 0.372 & \textbf{0.000} & \textbf{0.000} & \textbf{0.004} \\
\midrule
Mr\_Derivatives & BC & \underline{0.077} & 0.638 & 0.127 & 0.055 & \underline{0.225} & \underline{0.007} \\
 & IQL & 0.065 & 0.572 & 0.110 & 0.055 & 0.316 & 0.008 \\
 & CQL & 0.070 & \textbf{2.214} & \textbf{0.020} & 0.987 & 0.945 & 0.175 \\
 & TD3BC & \textbf{0.167} & \underline{1.854} & \underline{0.049} & 0.280 & 0.353 & 0.079 \\
 & AWAC & 0.071 & 0.621 & 0.111 & \underline{0.054} & 0.347 & 0.008 \\
 & Ours$^{\star}$ & -0.111 & -0.236 & 0.305 & \textbf{0.000} & \textbf{0.000} & \textbf{0.004} \\
\midrule
Stephanie\_Link & BC & -0.108 & -1.768 & 0.076 & 0.102 & 0.233 & 0.016 \\
 & IQL & -0.108 & -1.783 & 0.072 & \underline{0.094} & 0.261 & \underline{0.014} \\
 & CQL & \textbf{-0.093} & -2.202 & \underline{0.053} & 0.986 & 0.770 & 0.115 \\
 & TD3BC & \underline{-0.096} & -2.145 & \textbf{0.049} & 0.876 & 0.314 & 0.135 \\
 & AWAC & -0.101 & \underline{-1.629} & 0.068 & 0.095 & \underline{0.226} & 0.014 \\
 & Ours$^{\star}$ & -0.113 & \textbf{-0.234} & 0.289 & \textbf{0.000} & \textbf{0.000} & \textbf{0.004} \\
\midrule
Stocktwits & BC & 0.051 & 0.394 & 0.102 & 0.043 & 0.361 & 0.007 \\
 & IQL & 0.067 & 0.475 & 0.101 & 0.043 & 0.338 & 0.007 \\
 & CQL & -0.094 & -0.806 & \textbf{0.079} & 0.961 & \underline{0.243} & 0.080 \\
 & TD3BC & -0.145 & -1.325 & 0.165 & 0.355 & 0.556 & 0.028 \\
 & AWAC & \underline{0.088} & \underline{0.536} & \underline{0.100} & \underline{0.042} & 0.304 & \underline{0.006} \\
 & Ours$^{\star}$ & \textbf{0.255} & \textbf{0.762} & 0.294 & \textbf{0.000} & \textbf{0.000} & \textbf{0.002} \\
\midrule
traderstewie & BC & \textbf{0.005} & \underline{0.173} & \underline{0.061} & \underline{0.145} & \underline{0.240} & 0.021 \\
 & IQL & -0.006 & 0.017 & \textbf{0.060} & 0.151 & 0.269 & 0.021 \\
 & CQL & -0.046 & -1.048 & 0.087 & 0.908 & 0.579 & 0.100 \\
 & TD3BC & -0.052 & -0.785 & 0.106 & 0.979 & 0.953 & 0.240 \\
 & AWAC & \underline{-0.005} & 0.039 & 0.061 & 0.157 & 0.257 & \underline{0.021} \\
 & Ours$^{\star}$ & -0.083 & \textbf{0.322} & 0.376 & \textbf{0.000} & \textbf{0.000} & \textbf{0.005} \\
\bottomrule
\end{tabular}
}
\end{minipage}
\end{center}
\clearpage

\begin{center}
\centering
\captionof{table}{KOL-level results on YouTube.}
\label{tab:appendix_yt_kol_results}
\footnotesize
\setlength{\tabcolsep}{1.6pt}
\renewcommand{\arraystretch}{1.80}
\begin{minipage}[t]{0.49\textwidth}
\centering
\resizebox{\linewidth}{!}{
\begin{tabular}{>{\centering\arraybackslash}p{3.1cm} >{\centering\arraybackslash}p{1.35cm} cccccc}
\toprule
\textbf{KOL} & \textbf{Mth.} & \textbf{Ret.}$\uparrow$ & \textbf{Sha.}$\uparrow$ & \textbf{MDD}$\downarrow$ & \textbf{UER}$\downarrow$ & \textbf{DRR}$\downarrow$ & \textbf{BD}$\downarrow$ \\
\midrule
\shortstack{Ale\_s\_World\_of\\Stocks} & BC & 0.057 & 1.243 & 0.047 & \underline{0.101} & 0.376 & 0.015 \\
 & IQL & \underline{0.070} & 1.519 & 0.056 & 0.107 & 0.341 & \underline{0.015} \\
 & CQL & -0.050 & -1.746 & 0.061 & 0.933 & \underline{0.085} & 0.167 \\
 & TD3BC & 0.057 & \underline{2.475} & \textbf{0.013} & 0.991 & 0.856 & 0.197 \\
 & AWAC & 0.068 & 1.578 & \underline{0.042} & 0.109 & 0.362 & 0.015 \\
 & Ours$^{\star}$ & \textbf{0.491} & \textbf{2.503} & 0.125 & \textbf{0.000} & \textbf{0.000} & \textbf{0.006} \\
\midrule
Daniel\_Pronk & BC & -0.045 & -0.762 & \underline{0.062} & 0.257 & 0.209 & 0.039 \\
 & IQL & -0.018 & -0.161 & 0.069 & 0.259 & 0.180 & 0.037 \\
 & CQL & \textbf{0.052} & \textbf{2.453} & \textbf{0.028} & 0.968 & 0.398 & 0.307 \\
 & TD3BC & -0.069 & -3.884 & 0.076 & 0.948 & 0.896 & 0.252 \\
 & AWAC & -0.021 & -0.214 & 0.066 & \underline{0.236} & \underline{0.114} & \underline{0.034} \\
 & Ours$^{\star}$ & \underline{0.011} & \underline{0.405} & 0.089 & \textbf{0.000} & \textbf{0.000} & \textbf{0.013} \\
\midrule
Dividend\_Data & BC & 0.189 & \underline{6.694} & 0.015 & 0.103 & 0.265 & 0.013 \\
 & IQL & \underline{0.240} & \textbf{6.935} & \underline{0.014} & \underline{0.090} & 0.249 & 0.012 \\
 & CQL & -0.055 & -3.586 & 0.033 & 0.920 & 0.593 & 0.359 \\
 & TD3BC & 0.010 & 1.161 & \textbf{0.008} & 0.887 & 0.634 & 0.133 \\
 & AWAC & 0.233 & 6.434 & 0.022 & 0.091 & \underline{0.224} & \underline{0.012} \\
 & Ours$^{\star}$ & \textbf{0.456} & 4.966 & 0.074 & \textbf{0.000} & \textbf{0.000} & \textbf{0.005} \\
\midrule
Financial\_Education & BC & \underline{0.198} & \underline{3.152} & 0.109 & 0.146 & 0.256 & 0.021 \\
 & IQL & 0.191 & 2.737 & 0.109 & 0.145 & \underline{0.246} & 0.020 \\
 & CQL & 0.098 & \textbf{4.486} & \textbf{0.017} & 0.912 & 0.354 & 0.199 \\
 & TD3BC & -0.054 & -1.765 & \underline{0.104} & 0.963 & 0.566 & 0.146 \\
 & AWAC & 0.194 & 2.688 & 0.112 & \underline{0.143} & 0.256 & \underline{0.019} \\
 & Ours$^{\star}$ & \textbf{0.459} & 2.690 & 0.348 & \textbf{0.000} & \textbf{0.000} & \textbf{0.010} \\
\midrule
Humbled\_Trader & BC & 0.078 & 1.786 & 0.223 & 0.991 & 0.292 & 0.140 \\
 & IQL & 0.135 & 2.436 & 0.268 & \underline{0.743} & \textbf{0.000} & \underline{0.074} \\
 & CQL & 0.002 & 0.738 & 0.226 & 0.876 & 0.375 & 0.280 \\
 & TD3BC & 0.077 & 1.773 & \textbf{0.216} & 0.929 & \underline{0.042} & 0.524 \\
 & AWAC & \underline{0.194} & \underline{3.125} & \underline{0.220} & 0.929 & \textbf{0.000} & 0.119 \\
 & Ours$^{\star}$ & \textbf{0.353} & \textbf{4.225} & 0.285 & \textbf{0.000} & \textbf{0.000} & \textbf{0.012} \\
\bottomrule
\end{tabular}
}
\end{minipage}\hfill
\begin{minipage}[t]{0.49\textwidth}
\centering
\resizebox{\linewidth}{!}{
\begin{tabular}{>{\centering\arraybackslash}p{3.1cm} >{\centering\arraybackslash}p{1.35cm} cccccc}
\toprule
\textbf{KOL} & \textbf{Mth.} & \textbf{Ret.}$\uparrow$ & \textbf{Sha.}$\uparrow$ & \textbf{MDD}$\downarrow$ & \textbf{UER}$\downarrow$ & \textbf{DRR}$\downarrow$ & \textbf{BD}$\downarrow$ \\
\midrule
Invest\_with\_Henry & BC & 0.237 & 3.370 & 0.105 & \underline{0.201} & 0.132 & \underline{0.031} \\
 & IQL & \underline{0.305} & \underline{4.311} & 0.065 & 0.234 & \underline{0.110} & 0.031 \\
 & CQL & 0.164 & \textbf{5.285} & \textbf{0.013} & 0.937 & 0.205 & 0.199 \\
 & TD3BC & 0.006 & 0.319 & \underline{0.044} & 0.914 & 0.534 & 0.151 \\
 & AWAC & 0.270 & 3.738 & 0.083 & 0.226 & 0.135 & 0.031 \\
 & Ours$^{\star}$ & \textbf{0.370} & 3.324 & 0.092 & \textbf{0.000} & \textbf{0.000} & \textbf{0.013} \\
\midrule
Investing\_with\_Tom & BC & 0.001 & 0.149 & 0.032 & \underline{0.293} & 0.295 & \underline{0.040} \\
 & IQL & \underline{0.030} & 1.884 & 0.032 & 0.295 & \underline{0.192} & 0.042 \\
 & CQL & 0.018 & \underline{2.485} & \textbf{0.011} & 0.998 & \textbf{0.000} & 0.203 \\
 & TD3BC & 0.013 & 1.816 & \underline{0.016} & 1.000 & \textbf{0.000} & 0.232 \\
 & AWAC & 0.023 & 1.513 & 0.032 & 0.299 & 0.205 & 0.041 \\
 & Ours$^{\star}$ & \textbf{0.088} & \textbf{2.982} & 0.043 & \textbf{0.000} & \textbf{0.000} & \textbf{0.009} \\
\midrule
\shortstack{Sven\_Carlin\\Ph.D.} & BC & 0.058 & \underline{2.446} & \underline{0.024} & \underline{0.251} & 0.184 & \underline{0.034} \\
 & IQL & 0.052 & 2.201 & 0.029 & 0.288 & 0.175 & 0.035 \\
 & CQL & -0.087 & -5.917 & 0.087 & 0.864 & 0.563 & 0.193 \\
 & TD3BC & -0.105 & -3.202 & 0.105 & 0.960 & 0.728 & 0.164 \\
 & AWAC & \underline{0.086} & \textbf{3.215} & \textbf{0.021} & 0.289 & \underline{0.165} & 0.035 \\
 & Ours$^{\star}$ & \textbf{0.152} & 1.463 & 0.130 & \textbf{0.000} & \textbf{0.000} & \textbf{0.008} \\
\midrule
\shortstack{The\_Maverick\_of\\Wall\_Street} & BC & 0.122 & 1.603 & 0.084 & \underline{0.059} & 0.366 & 0.007 \\
 & IQL & \underline{0.210} & 2.358 & 0.074 & 0.060 & 0.378 & \underline{0.007} \\
 & CQL & 0.033 & \textbf{2.809} & \textbf{0.013} & 0.882 & 0.259 & 0.173 \\
 & TD3BC & 0.049 & \underline{2.663} & \underline{0.019} & 0.796 & \underline{0.053} & 0.084 \\
 & AWAC & 0.200 & 2.238 & 0.089 & 0.066 & 0.327 & 0.007 \\
 & Ours$^{\star}$ & \textbf{0.446} & 2.334 & 0.200 & \textbf{0.000} & \textbf{0.000} & \textbf{0.003} \\
\midrule
Unrivaled\_Investing & BC & -0.013 & -0.268 & 0.034 & 0.340 & \underline{0.017} & 0.037 \\
 & IQL & \underline{-0.002} & \underline{0.056} & \underline{0.029} & \underline{0.309} & 0.051 & \underline{0.036} \\
 & CQL & -0.050 & -1.891 & \textbf{0.013} & 0.938 & 0.475 & 0.294 \\
 & TD3BC & -0.039 & -1.020 & 0.043 & 0.961 & 0.729 & 0.173 \\
 & AWAC & -0.010 & -0.182 & 0.033 & 0.310 & 0.034 & 0.036 \\
 & Ours$^{\star}$ & \textbf{0.040} & \textbf{0.707} & 0.205 & \textbf{0.000} & \textbf{0.000} & \textbf{0.006} \\
\bottomrule
\end{tabular}
}
\end{minipage}
\end{center}

\normalsize
\renewcommand{\arraystretch}{1.0}

\clearpage

\section{Daily Equity Curves}
\label{app:daily_equity_curves}

We report main quantitative results at the event level, but visualize equity curves at the daily mark-to-market level. Event-level metrics are used in the main tables because both training samples and policy decisions are event-conditioned, while daily curves provide complementary evidence on continuous portfolio evolution between decision events.

\begin{figure}[!htbp]
\centering
\includegraphics[width=\textwidth,height=0.76\textheight,keepaspectratio]{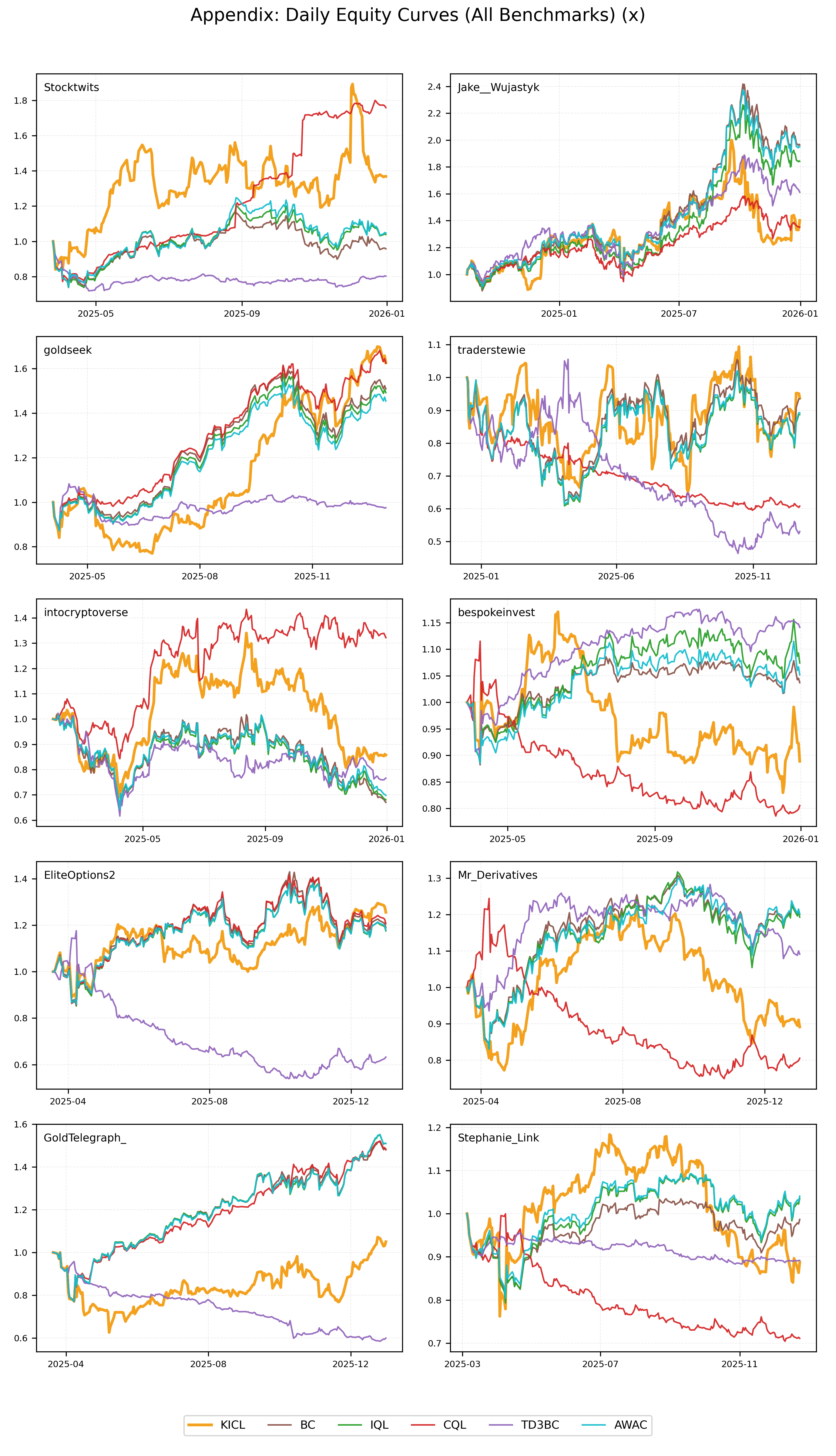}
\caption{Daily mark-to-market equity curves on the selected X benchmark subset.}
\label{fig:daily_equity_x}
\end{figure}

\clearpage

\begin{figure}[!htbp]
\centering
\includegraphics[width=\textwidth,height=0.76\textheight,keepaspectratio]{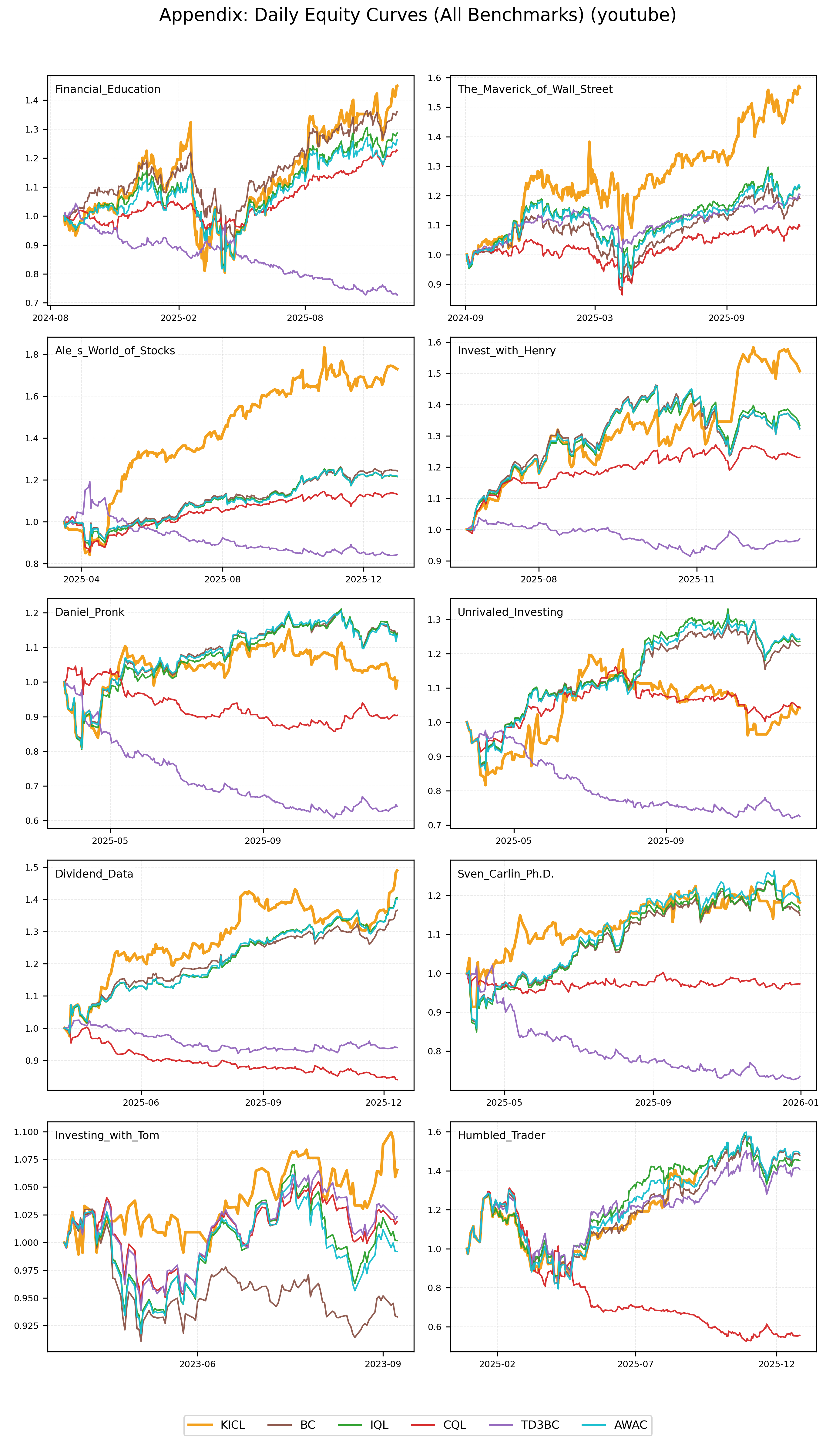}
\caption{Daily mark-to-market equity curves on the selected YouTube benchmark subset.}
\label{fig:daily_equity_youtube}
\end{figure}

\clearpage

\end{document}